\documentclass[12pt,draftclsnofoot,journal,onecolumn]{IEEEtran}
\usepackage{tabularx}
\usepackage[T1]{fontenc}
\usepackage[noend]{algpseudocode}
\usepackage{graphicx,epstopdf}
\usepackage{amsmath,amssymb,amsfonts,stfloats}
\usepackage{array}
\usepackage{textcomp}
\usepackage[font=small]{caption}
\usepackage{paralist}
\usepackage{lettrine}
\usepackage{cite}
\usepackage{xcolor}
\usepackage{cancel}
\usepackage{booktabs}
\usepackage[export]{adjustbox}
\usepackage[caption=false,font=footnotesize]{subfig}
\usepackage{url}
\usepackage{acronym}
\usepackage{multicol}
\usepackage[ruled,vlined]{algorithm2e}
\newcommand{\norm}[1]{\left\lVert#1\right\rVert}

\begin{document}

\title{Deep Learning of Transferable MIMO Channel Modes for 6G V2X Communications}

\author{Lorenzo~Cazzella, Dario~Tagliaferri, Marouan~Mizmizi, Damiano~Badini, Christian~Mazzucco, Matteo~Matteucci, and~Umberto~Spagnolini}

\maketitle

\begin{abstract}
In the emerging high mobility Vehicle-to-Everything (V2X) communications using millimeter Wave (mmWave) and sub-THz, Multiple-Input Multiple-Output (MIMO) channel estimation is an extremely challenging task. At mmWaves/sub-THz frequencies, MIMO channels exhibit few leading paths in the space-time domain (i.e., directions or arrival/departure and delays). Algebraic Low-rank (LR) channel estimation exploits space-time channel sparsity through the computation of \textit{position-dependent} MIMO channel eigenmodes leveraging recurrent training vehicle passages in the coverage cell. LR requires vehicles’ geographical positions and tens to hundreds of training vehicles’ passages for \textit{each} position, leading to significant complexity and control signalling overhead. Here we design a DL-based LR channel estimation method to infer MIMO channel eigenmodes in V2X urban settings, starting from a single LS channel estimate and without needing vehicle's position information. Numerical results show that the proposed method attains comparable Mean Squared Error (MSE) performance as the position-based LR. Moreover, we show that the proposed model can be trained on a reference scenario and be effectively transferred to urban contexts with different space-time channel features, providing comparable MSE performance without an explicit transfer learning procedure. This result eases the deployment in arbitrary dense urban scenarios.
\end{abstract}

\begin{IEEEkeywords}
   MIMO, Deep learning, Channel estimation, V2X, Millimeter-wave, sub-THz, 6G
\end{IEEEkeywords}

\IEEEpeerreviewmaketitle

\section{Introduction}
Millimeter Wave (mmWave) ($30-100$ GHz) and sub-THz ($100-300$ GHz) bands arose as the leading solution to overcome the bandwidth scarcity occurring in the sub-6 GHz EM spectrum, e.g., $0.41-7.125$ GHz in 5G New Radio (NR) Frequency Range 1 (FR1). In particular, mmWaves in the $24.25-52.6$ GHz range are designated for 5G NR FR2~\cite{Garcia2021_5GNRtutorial}, while sub-THz W- and D-bands will be the pillars of 6G paradigm by 2030, to accommodate the increasing capacity requirements such as for Vehicle-to-Everything (V2X)-enabled services~\cite{Wymeersch2021_6G}. By increasing the carrier frequency, the propagation is affected by an orders-of-magnitude increase in the path-loss, inducing coverage reduction in Non Line-Of-Sight (NLOS) scenarios and a \textit{sparse} communication channel characterized by few significant paths in the Space-Time (ST) domain of Directions of Arrival/Departure (DoAs/DoDs) and delays~\cite{6834753,Akyildiz2015_subTHz,wang2018survey}. In this regard, massive Multiple-Input Multiple-Output (MIMO) systems, enabled by reduced antenna footprints at mmWave and sub-THz, are used to counteract the path-loss by beamforming strategies at both Transmitter (Tx) and Receiver (Rx)~\cite{kutty2015beamforming}.

In massive MIMO systems, the channel knowledge is essential for designing the correct Tx and Rx beamforming. Legacy multi-carrier systems, such as Orthogonal Frequency Division Multiplexing/Multiple Access (OFDM/OFDMA) 5G NR FR2 radio interface systems, leverage Least Squares (LS) MIMO channel estimation from known pilot sequences~\cite{TS38213}. LS channel estimation is known to be inaccurate in low Signal-to-Noise Ratio (SNR) conditions and large MIMO settings, when the number of unknowns increases with the number of antennas and the bandwidth. LS has been improved by exploiting the sparsity of the mmWave/sub-THz MIMO channel. Structured methods aim at directly estimating the physical channel ST features such as DoAs/DoDs/delays, using either super-resolution methods, such as MUltiple SIgnal Classification (MUSIC)~\cite{Guo2017_MUSIC_mmWChEst}, Estimation of Signal Parameters via Rotational Invariance Technique (ESPRIT)~\cite{Liao2017_ESPRIT_ChEst}, or by constraining the sparsity in a suitable optimization problem, as in Compressed Sensing (CS)~\cite{bajwa2010compressed}. These approaches allow to estimate the MIMO channel with high accuracy at the price of being sensitive to hardware impairments (i.e., antenna array calibration and coupling theory)~\cite{Wang2018_CShardwareimpairments}. 

An alternative to a structured channel estimation is based on algebraic theory. Algebraic Low-Rank (LR) methods combine high accuracy with an inherent robustness to hardware impairments~\cite{Nicoli2003,Cerutti2020}. LR operate on multiple pilot sequences transmitted from a single (or multiple) collaborative User Equipment (UE) and collected by a fixed Base Station (BS), where each pilot transmission shares the same DoAs, DoDs and delays, while single paths' fading amplitudes are assumed to vary according to the Doppler spectrum. Indeed, the ensemble of received pilot sequences are used to compute the spatial and temporal \textit{modes} of the MIMO channel to filter new pilot signals to retrieve the LR channel estimate \cite{Cerutti2020}.

From an algebraic point of view, LR only requires the \textit{stationarity of the ST channel eigenmodes}, and there is no need to explicitly estimate DoAs, DoDs and delays, as the channel modes are \textit{unstructured}, resulting more robust to antenna calibration issues. The LR efficacy is proportional to the sparsity of the MIMO channel matrix. Early works on LR were targeted to sub-6 GHz systems~\cite{Nicoli2003}, while more recent ones were tailored for mmWave and compared with CS~\cite{Cerutti2020}. LR channel estimation \cite{Cerutti2020} leverages hundreds or thousands consecutive transmissions from the same moving collaborative UE towards the BS, limiting the application to static or quasi-static communication scenarios.

\begin{figure}[!t]
    \centering
    \subfloat[][]{\includegraphics[width=.45\textwidth]{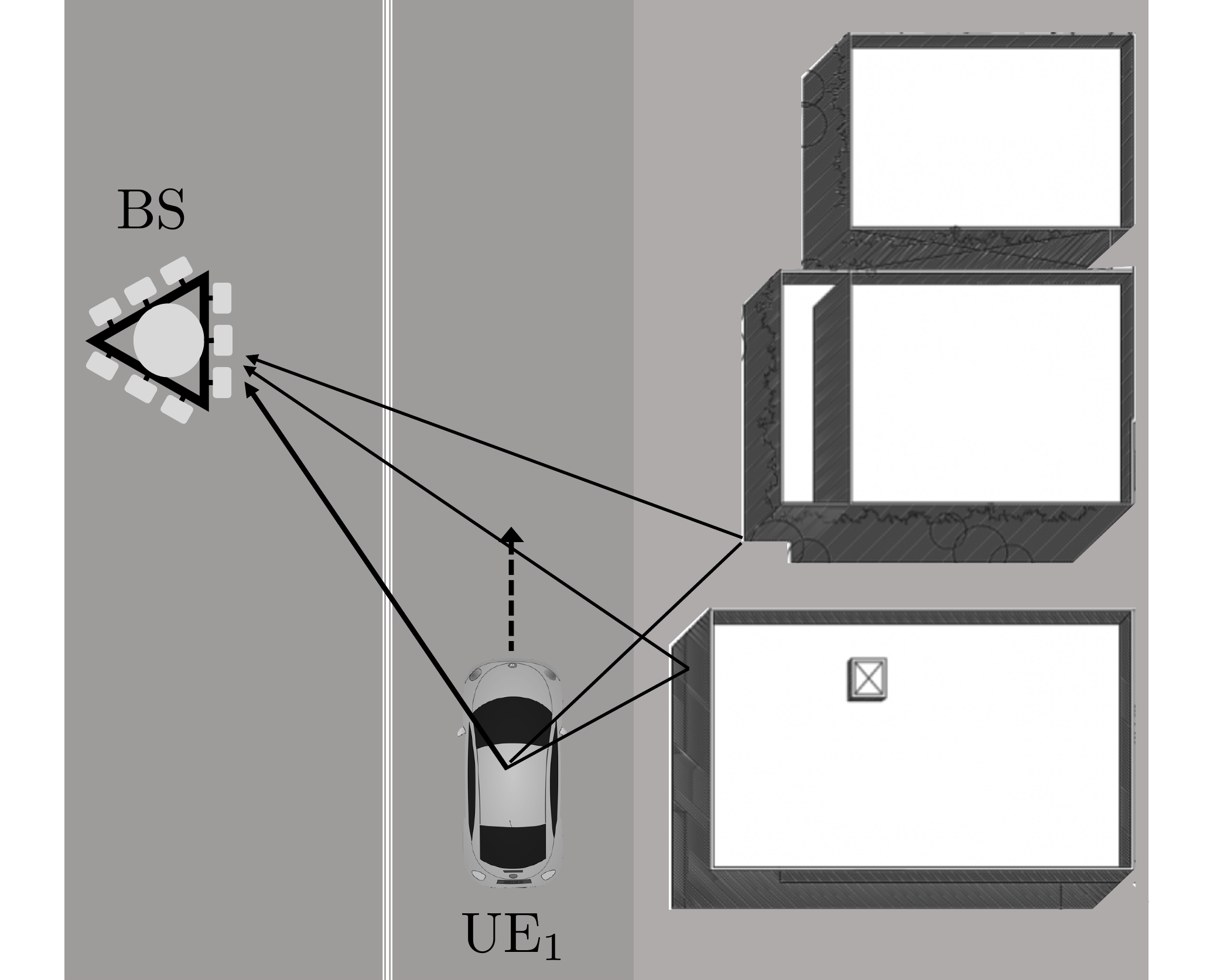}}
    \subfloat[][]{\includegraphics[width=.45\textwidth]{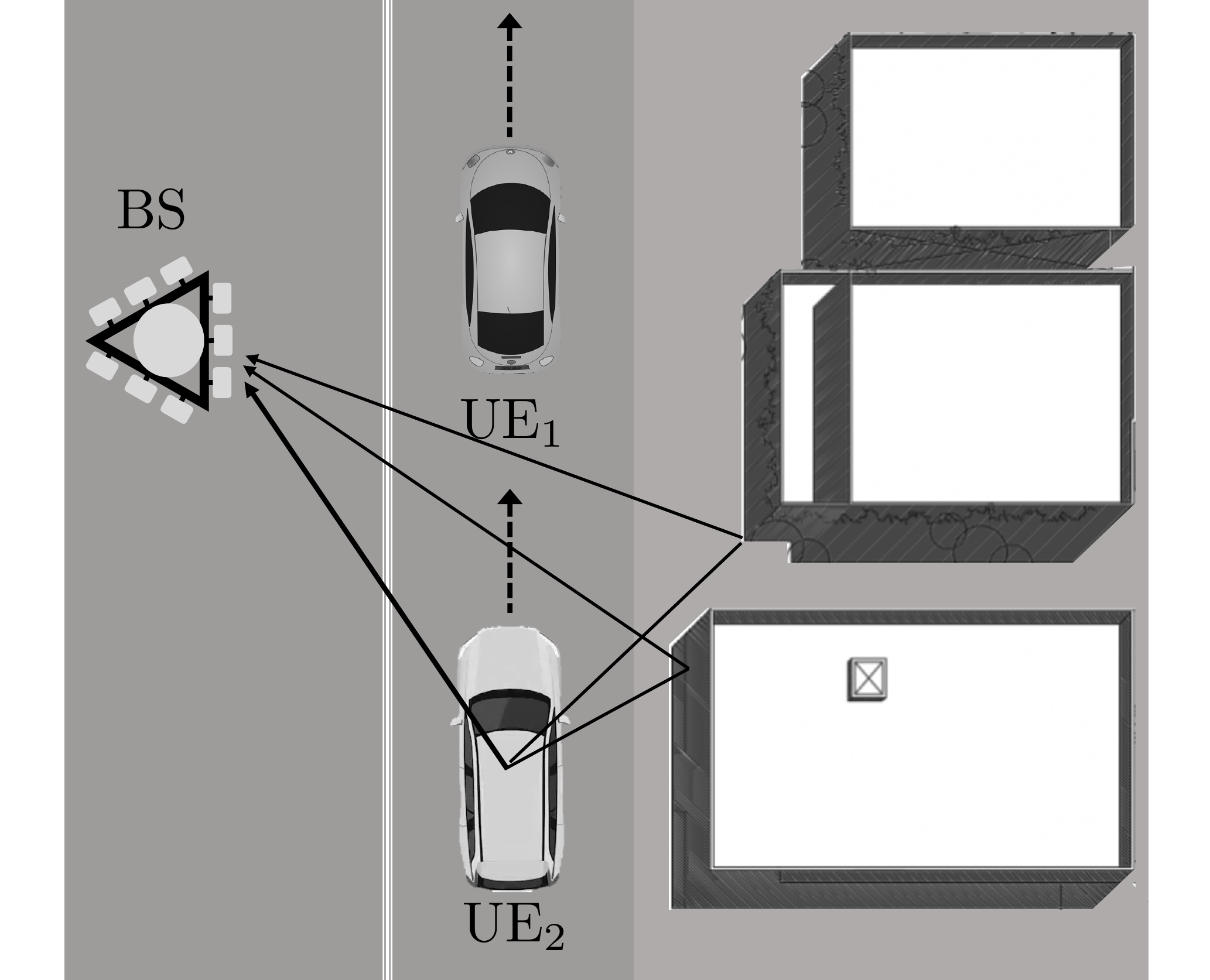}}\\
    \caption{Vehicular MIMO channel recurrences induced by road constraints: UE$_1$ and UE$_2$ experience the same DoDs, DoAs and delays in communicating with the BS when passing over the same location in the cell.}
    \label{fig:recurrences}
\end{figure}

In our previous work~\cite{mizmizi2021channel}, we overcome this limitation by collecting the set of received pilot sequences on \textit{recurrent vehicle passages} over the same geographical area, to ensure the same ST channel structure for each received sequence. The key idea is that roads constrain vehicles to have recurrent passages and thus the associated MIMO channels share similar ST channel structures over different vehicles, as depicted in Fig. \ref{fig:recurrences}. The LR channel modes are thus related to physical UEs' positions in the cell, and this is suitable for V2X systems. Still, the method presented in~\cite{mizmizi2021channel} requires the availability of a suitable number (tens to hundreds) of collaborative UEs, i.e., vehicles transmitting their position, for \textit{each} position within a given coverage cell. When the number of cells grows, the complexity of a position-based LR method rapidly becomes overwhelming. Furthermore, LR requires the continuous exchange of UEs position information, imposing a non-negligible BS-UE signalling.

Deep-Learning (DL) is foreseen to play a pivotal role in 6G, complementing or even substituting standard tasks introduced by novel communication frameworks, as massive MIMO systems at mmWave/sub-THz frequencies or reconfigurable intelligent surfaces \cite{SalhDL6G2021}, increasing the adaptability of the communication system to the local conditions of the environment. DL learns complex tasks from data where model-based techniques fail or turn out to be sub-optimal, exploiting Deep Neural Networks (DNNs)~\cite{osheaIntroductionDeepLearning2017a,zhangDeepLearningMobile2019,huangDeepLearningPhysicalLayer2020}. 
Recently, many works have addressed the problem of MIMO channel estimation by means of DL models. A seminal work in this direction is \cite{osheaIntroductionDeepLearning2017a}, which introduces the application of DL techniques at the physical layer. A DNN with a convolutional denoiser was derived from the \textit{learned denoising-based approximate message passing} algorithm \cite{wen2018deep}. Channel estimation for fast time-varying MIMO OFDM systems in mobility is based on convolutional long short-term memory NN in \cite{liao2019deep}. Exploiting the deep image prior framework \cite{ulyanov2018deep}, the work \cite{balevi2020massive} proposes instead a massive MIMO channel estimation method through an untrained deep neural network. 
Transfer Learning (TL) has been also recently considered as a powerful tool to extend and transfer the knowledge from one task to another that shares some inherent commonalities, by re-training only a subset of the DNN layers \cite{nguyen2021transfer}. On channel estimation, a deep TL method exploits previously trained models to accelerate site adaptation \cite{alves2021deep}. The downlink channel prediction is addressed in \cite{yang2020deep} as a deep TL problem, proposing the use of fully-connected neural network architectures and fine-tuning trained models for new environments. 

To advance with respect to current state of the art, in this paper we propose the following contributions:  
\begin{itemize}
    \item We propose a DL-based approach to infer the ST MIMO channel eigenmodes of LR channel estimation for 6G V2X with high-mobility. Compared to the reference position-based LR approach~\cite{mizmizi2021channel}, advantages (after the initial training of the DNN) are the lack of UE position information at BS, and associated control signalling. Simulations by ray-tracing (to generate channel data)~\cite{altairwinprop} over realistic vehicle trajectories~\cite{SUMO2018} prove feasibility and benefits of the proposed DL-based LR approach, that outperforms the conventional LS estimation in terms of Mean Squared Error (MSE) by more than one order of magnitude ($\approx 15$ dB on average). MSE performance of DL-based LR is comparable with position-based LR method for both frequency-flat and frequency-selective MIMO channels. Notice that MSE performance of the position-based LR method reaches the theoretical lower MSE bound in \cite{Nicoli2003}, thus our method is statistically efficient too.
    
    \item We show that the proposed DL-based LR channel estimation model generalizes over different urban scenarios, each characterized by different ST channel eigenmodes. Simulation results indicate that it is possible to perform the DNN training over a single scenario (exploiting pilot symbols from collaborative UEs in the reference position-based LR method) and transfer the learned algebraic MIMO channel structure to other scenarios, still outperforming LS. In particular, for frequency-flat MIMO channels, there is no practical advantage in employing additional TL procedures. For frequency-selective channels, the explicit re-training of the last 2 fully-connected layers of the DNN reduces the average MSE by $\approx 2 $ dB (consistently over 5 different scenarios). 
    
\end{itemize}

We remark that the proposed approach is substantially different from the existing ones \cite{wen2018deep,liao2019deep,balevi2020massive}, which are targeted to learning either the physical (structured) channel features or directly the MIMO channel matrix entries. The advantage of the proposed approach is indeed the robustness against hardware impairments, inherited from the LR MIMO channel estimation \cite{Cerutti2020}. Moreover, as shown in our previous work \cite{cazzella2021positionagnostic}, MIMO channel eigenmodes can be effectively grouped in few ($<10$) clusters in space, much less that the possible MIMO channel configurations. This characteristic eases the information transfer from one scenario to another, reducing the overall number of collaborative vehicular UEs used for DNN training to a single reference scenario.

The paper is organized as follows: Section \ref{sect:system&channelmodel} outlines the analytical system and channel model used in this paper; Section \ref{sect:LR} summarizes the reference position-based LR channel estimation method, functionally to the application of the DL technique in Section \ref{sect:DL}; Section \ref{sect:results} reports the simulation results while Section \ref{sect:conclusion} draws some final conclusions.

\subsection*{Notation}
Bold upper- and lower-case letters describe matrices and column vectors. Matrix transposition, conjugation, conjugate transposition and Frobenius norm are indicated respectively as $\mathbf{A}^{\mathrm{T}}$, $\mathbf{A}^{*}$, $\mathbf{A}^{\mathrm{H}}$ and $\norm{\mathbf{A}}$. $\mathrm{tr}\left(\mathbf{A}\right)$, $\mathrm{rank}\left(\mathbf{A}\right)$ extract, respectively, the trace and the rank of $\mathbf{A}$. Symbol $\otimes$ denotes the Kronecker product between two matrices. $\mathrm{vec}(\mathbf{A})$ denotes the vectorization by columns of $\mathbf{A}$. $\mathrm{diag}(\mathbf{A})$ denotes the extraction of the diagonal of $\mathbf{A}$, while $\mathrm{diag}(\mathbf{a})$ is the diagonal matrix given by vector $\mathbf{a}$. $\mathbf{I}_n$ is the identity matrix of size $n$. The Cholesky decomposition of a positive-definite matrix $\mathbf{A}$ is $\mathbf{A}=\mathbf{A}^{\frac{\mathrm{H}}{2}}\mathbf{A}^{\frac{1}{2}}$, where $\mathbf{A}^{\frac{\mathrm{H}}{2}}$ is the lower-triangular unique square root of $\mathbf{A}$. The following property of the vectorization is used in the text: $\mathrm{vec}(\mathbf{A}\mathbf{B}) = (\mathbf{B}^{\mathrm{T}} \otimes \mathbf{I})\mathrm{vec}(\mathbf{A})$. With  $\mathbf{a}\sim\mathcal{CN}(\boldsymbol{\mu},\mathbf{C})$ we denote a multi-variate circularly complex Gaussian random variable $\mathbf{a}$ with mean $\boldsymbol{\mu}$ and covariance $\mathbf{C}$. $\mathbb{E}[\cdot]$ is the expectation operator, while $\mathbb{R}$ and $\mathbb{C}$ stand for the set of real and complex numbers, respectively. $\delta_{n}$ is the Kronecker delta.

\section{System and Channel Model}\label{sect:system&channelmodel}

We consider a single-user, multi-carrier uplink communication system over a bandwidth $B$, in which the Tx and the Rx are equipped with $N_T$ and $N_R$ antennas.
At the receiving antennas, after the time and frequency synchronization and cyclic prefix removal, the Rx signal is:
\begin{equation}\label{eq:Rxsignal}
    \mathbf{y}(t) = \mathbf{H}(t) * \mathbf{x}(t) + \mathbf{n}(t),
\end{equation}
where symbol $*$ denotes the matrix convolution between the transmitted signal $\mathbf{x}(t)=[x_1(t),\dots, x_{N_T}(t)]^{\mathrm{T}}\in\mathbb{C}^{N_T \times 1}$ at each Tx antenna and the $N_R N_T$ MIMO channel responses
\begin{equation}
    \mathbf{H}(t) = \begin{bmatrix} h_{11}(t) &\cdots & h_{1 N_T}(t)\\
    h_{21}(t) &\cdots & h_{2 N_T}(t)\\
    \vdots & \vdots & \vdots\\
    h_{N_R 1}(t) &\cdots & h_{N_R N_T}(t)\\
    \end{bmatrix}\in\mathbb{C}^{N_R \times N_T},
\end{equation}
where $h_{nm}(t)$ is the impulse response from the $m$-th Tx antenna to the $n$-th Rx antenna, whose maximum temporal support of the MIMO channel is limited to $\tau_{max}$, $\forall n,m$. Vector $\mathbf{n}(t)\in\mathbb{C}^{N_R \times 1}$ denotes the additive Gaussian disturbance corrupting the received signal, comprising thermal noise and interference. By sampling \eqref{eq:Rxsignal} at time $t=wT$, where $T=1/B$, we obtain the discrete-time signal 
\begin{equation}\label{eq:Rxsignal_discrete}
    \mathbf{y}[w] = \mathbf{H}[w] * \mathbf{x}[w] + \mathbf{n}[w], 
\end{equation}
for $w=0,\dots,W-1$, where $W=\lceil \tau_{max}/T\rceil$ is the maximum number of channel taps and $\mathbf{H}[w]\equiv \mathbf{H}(wT)$ is the discrete-time MIMO channel matrix. For channel estimation purposes, the Tx signal $\mathbf{x}[w]$ is modelled as a random pilot sequence (known at the Rx), uncorrelated in time and space, i.e., $\mathbb{E}\left[\mathbf{x}[w] \mathbf{x}[\ell]^{\mathrm{H}}\right] = \sigma^2_x \mathbf{I}_{N_T} \delta_{w-\ell}$ ($\sigma^2_x$ is the signal power). The noise $\mathbf{n}[w]\sim\mathcal{CN}(\mathbf{0},\mathbf{Q}_n)$ is instead white in time but generally correlated in space, to account for directional interference, as
$\mathbb{E}\left[\mathbf{n}[w] \mathbf{n}[\ell]^{\mathrm{H}}\right] = \mathbf{Q}_n \delta_{w-\ell}$. 
The SNR measured at each antenna is:
\begin{equation}
    \mathrm{SNR} = \frac{\mathbb{E}\left[\big\|\sum_{w} \mathbf{H}[w]* \mathbf{x}[w]\big\|^2\right]}{\mathrm{tr}(\mathbf{Q}_n)}.
\end{equation}

In the following, we detail the analytical model for the MIMO channel discrete impulse response $\mathbf{H}[w]$, to better clarify the application of the DL-based LR channel estimation proposed in Section \ref{sect:DL}.

\subsection{MIMO Channel Model}
The mmWave/sub-THz MIMO channel impulse response is modelled as the sum of $P$ paths as~\cite{6834753}:
\begin{equation}\label{eq:channel_matrix_time}
    \mathbf{H}(t) = \sum_{p=1}^{P}\beta_p \, e^{j 2 \pi \nu_p t}\,\mathbf{a}_R(\boldsymbol{\theta}_p)\mathbf{a}_T^{\mathrm{T}}(\boldsymbol{\phi}_p)\,g(t-\tau_p),
\end{equation}
where the $p$-th path amplitude $\beta_p$ depends on path-loss and propagation geometry; $\nu_p$ is the $p$-th path Doppler shift; $\mathbf{a}_T(\boldsymbol{\phi}_p)\in\mathbb{C}^{N_T \times 1}$ and $\mathbf{a}_R(\boldsymbol{\theta}_p)\in\mathbb{C}^{N_R \times 1}$ are the Tx and Rx array response vectors to the $p$-th path, respectively, function of the DoDs $\boldsymbol{\phi}_p = [\phi^{\mathrm{az}}_p, \phi^{\mathrm{el}}_p]^{\mathrm{T}}$ and the DoAs $\boldsymbol{\theta}_p = [\theta^{\mathrm{az}}_p, \theta^{\mathrm{el}}_p]^{\mathrm{T}}$ (for azimuth and elevation); $g(t-\tau_p)$ is the pulse shaping waveform (typically a raised cosine) delayed by $\tau_p$ ($p$-th path delay). Without loss of generality, we consider half-wavelength spaced uniform planar arrays with isotropic antennas for both Tx and Rx. The Tx array response is structured as:
\begin{equation}\label{eq:Txarrayresponse}
    \mathbf{a}_T(\boldsymbol{\phi}_p) = \mathbf{a}^{\mathrm{el}}_T(\phi^{\mathrm{el}}_p) \otimes \mathbf{a}^{\mathrm{az}}_T(\phi^{\mathrm{az}}_p),
\end{equation}
where $\mathbf{a}^{\mathrm{az}}_T(\phi^{\mathrm{az}}_p)=[1,\dots,e^{j\pi (N_T-1) \sin(\phi^{\mathrm{az}}_p)}]$ and $\mathbf{a}^{\mathrm{el}}_T(\phi^{\mathrm{el}}_p)=[1,\dots,e^{j\pi (N_T-1) \sin(\phi^{\mathrm{el}}_p)}]$ are the steering vectors along azimuth and elevation DoDs. The Rx steering vector $\mathbf{a}_R(\boldsymbol{\theta}_p)$ is similarly structured.
We also assume that the Doppler-related rotation is almost constant over $\tau_{\mathrm{max}}$ (normalized to the first echo), such that $\alpha_p = \beta_p \, e^{j 2 \pi \nu_p t}\sim \mathcal{CN}\left(0,\Omega_p\right)$, obeying the wide-sense stationary uncorrelated scattering model. The latter implies the uncorrelation between any pair of scattering amplitudes $\mathbb{E}\left[ \alpha_{p,\ell}^{} \alpha_{q,k}^*\right] = \Omega_p \delta_{p-q}\delta_{\ell-k}$, where $\alpha_{p,\ell}$ is the scattering amplitude of the $p$-th path of the $\ell$-th channel.

By sampling \eqref{eq:channel_matrix_time} at $t=wT$ we obtain a compact matrix formulation of the MIMO channel
\begin{equation}\label{eq:channel_matrix_tap_time}
\begin{split}
        \mathbf{H}[w] &= \sum_{p=1}^{P}\alpha_p \,\mathbf{a}_R(\boldsymbol{\theta}_p)\mathbf{a}_T^{\mathrm{T}}(\boldsymbol{\phi}_p)\,g\left[wT-\tau_p\right] = \\
        & = \mathbf{A}_R\left(\boldsymbol{\theta}\right)
        \boldsymbol{\Lambda}[w]\mathbf{A}_T^{\mathrm{T}}\left(\boldsymbol{\phi}\right), \;\; w=0,\dots,W-1
\end{split}
\end{equation}
where $\mathbf{A}_T\left(\boldsymbol{\phi}\right) = \left[\mathbf{a}_T(\boldsymbol{\phi}_1),\dots,\mathbf{a}_T(\boldsymbol{\phi}_P)\right] \in \mathbb{C}^{N_T\times P}$ and
$\mathbf{A}_R\left(\boldsymbol{\theta}\right) = \left[\mathbf{a}_R(\boldsymbol{\theta}_1),\dots,\mathbf{a}_R(\boldsymbol{\theta}_P)\right] \in \mathbb{C}^{N_R\times P}$  are two frequency-independent matrices embedding the spatial channel features, and $\boldsymbol{\Lambda}[w] = \mathrm{diag}(\alpha_1 \, g[wT-\tau_1],\dots,\alpha_P\, g[wT-\tau_P]) \in \mathbb{C}^{P\times P}$ is a diagonal matrix collecting all the channel amplitudes scaled by the $w$-th tap of the pulse shaping waveform. 

Algebraic analysis of the matrixes $\mathbf{A}_T\left(\boldsymbol{\phi}\right)$ and $\mathbf{A}_R\left(\boldsymbol{\theta}\right)$ defines the spatial diversity orders of the MIMO channel in terms of the number of distinguishable rays at Tx and Rx, given the number of antennas $N_T$ and $N_R$. The diversity orders are expressed as
\begin{align}\label{eq:TX_RX_diversity_orders}
    r_{\mathrm{S}}^{\mathrm{Tx}} & = \mathrm{rank}(\mathbf{A}_T\left(\boldsymbol{\phi}\right)) \leq \mathrm{min}\left(N_T, P\right)\\
    r_{\mathrm{S}}^{\mathrm{Rx}} & = \mathrm{rank}(\mathbf{A}_R\left(\boldsymbol{\theta}\right))  \leq \mathrm{min}\left(N_R, P\right)
\end{align}
for Tx and Rx, respectively. Orders are limited by either the number of channel paths or by the number of antennas. Usually, mmWave and sub-THz channels are characterized by $P < N_T,N_R$. 

To ease the analytical derivations in Section \ref{sect:LR} and the application of DL in Section \ref{sect:DL}, we can further manipulate \eqref{eq:channel_matrix_tap_time} to extract the temporal (delays-related) diversity order of the MIMO channel as:
\begin{equation}\label{eq:channel_ST_matrix}
    \boldsymbol{\mathcal{H}} =\boldsymbol{\mathcal{A}}\left(\boldsymbol{\theta},\boldsymbol{\phi}\right) \mathbf{D} \,\mathbf{G}^{\mathrm{T}}(\boldsymbol{\tau}),
\end{equation}
where: $\boldsymbol{\mathcal{H}} = [\mathrm{vec}(\mathbf{H}[0]),\dots,\mathrm{vec}(\mathbf{H}[W-1])]\in\mathbb{C}^{N_T N_R \times W}$ is the ST channel matrix, whose Least Squares estimate is used as input to the DNN proposed in Section \ref{sect:DL}; $\boldsymbol{\mathcal{A}}\left(\boldsymbol{\theta},\boldsymbol{\phi}\right) = [\mathbf{a}_T(\boldsymbol{\phi}_1)\otimes\mathbf{a}_R(\boldsymbol{\theta}_1),\dots,\mathbf{a}_T(\boldsymbol{\phi}_P)\otimes\mathbf{a}_R(\boldsymbol{\theta}_P)]$ comprises both the DoDs and DoAs; $\mathbf{D} = \mathrm{diag}(\alpha_1,\dots,\alpha_P)$, and matrix $\mathbf{G}\left(\boldsymbol{\tau}\right) = \left[\mathbf{g}(\tau_1),\dots,\mathbf{g}(\tau_P)\right]$ embeds the temporal features $\boldsymbol{\tau}=[\tau_1,\dots,\tau_P]$ through vectors $\mathbf{g}\left(\tau_p\right) \in\mathbb{R}^{W \times 1} = \left[g\left[-\tau_p\right],\dots,g\left[(W-1)T-\tau_p\right]\right]^{\mathrm{T}}$. 
\begin{figure}[!t]
    \centering
    \includegraphics[width=.5\textwidth]{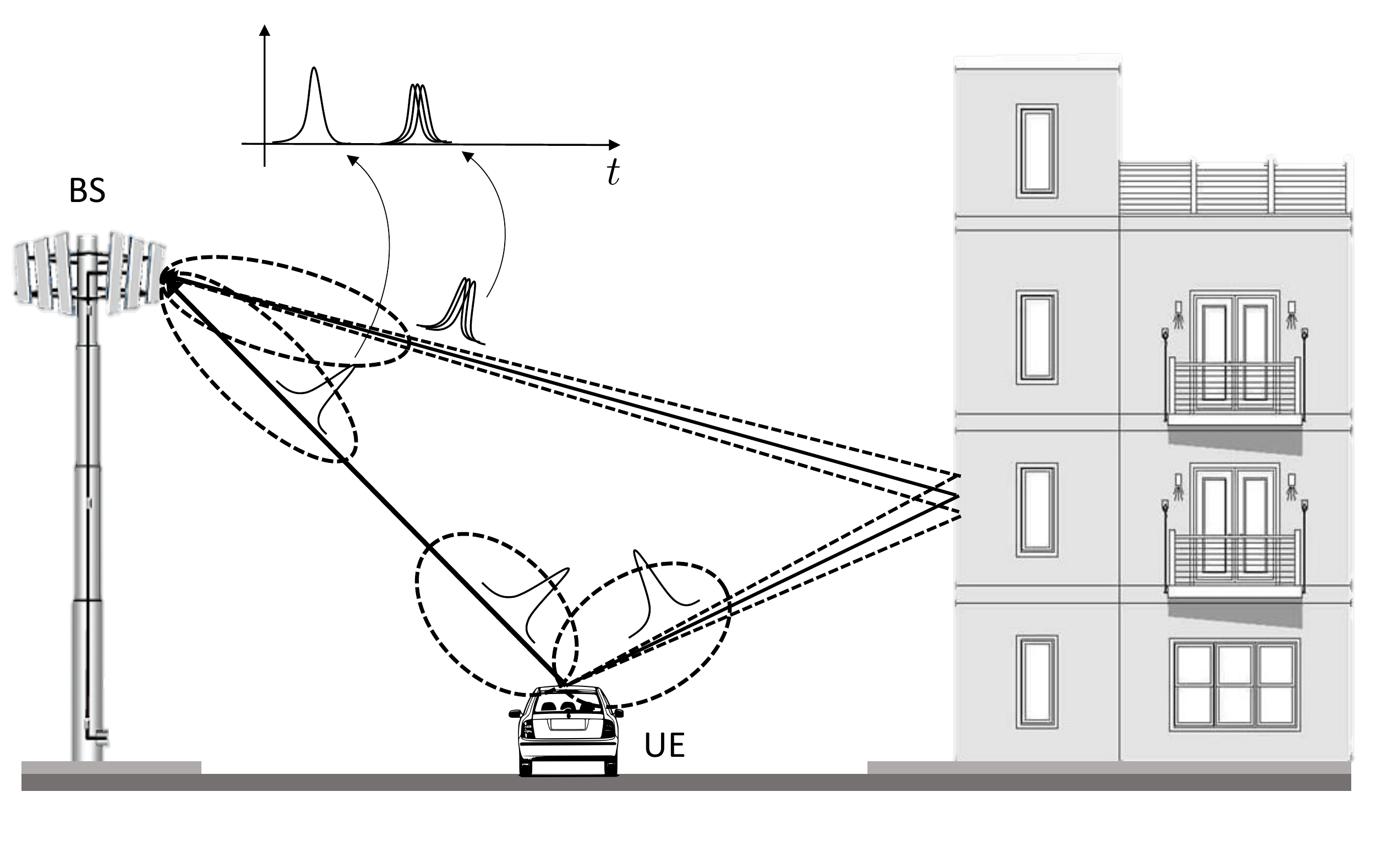}
    \caption{Effect of Tx/Rx spatial and temporal selectivity on a multipath scenario ($P=4$): reflections represented with dashed lines are spatially and temporally indistinguishable from the solid line one, due to the Tx and Rx beamwidths and bandwidth, therefore $r_{\mathrm{S}}^{\mathrm{Tx}} = r_{\mathrm{S}}^{\mathrm{Rx}} = r_{\mathrm{T}} = 2$.}
    \label{fig:ranks}
\end{figure}
The temporal diversity order is therefore:
\begin{align}\label{eq:S_T_diversity_orders}
    r_{\mathrm{T}} = \mathrm{rank}(\mathbf{G}\left(\boldsymbol{\tau}\right))  \leq \mathrm{min}\left(W, P\right),
\end{align}
ruled by the pulse width $T$ (and therefore by bandwidth $B$). The meaning of spatial and temporal channel orders is depicted in Fig. \ref{fig:ranks}, while the different channel manipulations used throughout the paper are reported in Table \ref{tab:channel_arrangments}.

\begin{table}[t!]
\begin{center}
\caption{Channel manipulations}
\begin{tabular}{lcl}
\toprule
\textbf{Symbol} & \textbf{Dimensions} & \textbf{Description}  \\
\noalign{\smallskip}
\hline
\noalign{\smallskip}
$\mathbf{h}$ & $W N_T N_R \times 1$ & time-space(Tx)-space(Rx) vector \\\noalign{\smallskip}
$\boldsymbol{\mathcal{H}}$ & $N_T N_R \times W$ & space(Tx+Rx)-time matrix \\\noalign{\smallskip}
$\mathbf{H}[w]$ & $N_R \times N_T$ & space(Rx)-space(Tx) matrix ($w$-th sample) \\\noalign{\smallskip}
\bottomrule
\end{tabular}
\label{tab:channel_arrangments}
\end{center}
\end{table}

\section{Position-based LR Channel Estimation}\label{sect:LR}

This section reports the algebraic background for the LR channel estimation leveraging $L$ different received pilot sequences $\{\mathbf{y}_{\ell}[w]\}_{\ell=1}^{\ell=L}$, assumed to be collected by the BS from \textit{different} vehicular UEs passing in the same location within the cell. Each UE is also requested to share with the BS its geographical position, obtained from on-board sensors or other techniques \cite{8690640}. Thus, sequences $\{\mathbf{y}_{\ell}[w]\}_{\ell=1}^{\ell=L}$ share the same ST propagation pattern. The complete analytical treatment, beyond the scope of the present work, can be found in \cite{Nicoli2003}.
In brief, the LR-estimated channel is retrieved through the application of a \textit{pilot-specific} matrix $\mathbf{T}_{\ell}$, providing the conventional LS MIMO channel estimate, and a \textit{position-specific} matrix $\boldsymbol{\Pi}_L(\bar{p})$ ($\bar{p}$ denotes a given position in the radio cell) on a single received pilot signal $\mathbf{y}_{\ell} = \left[\mathbf{y}^{\mathrm{T}}_{\ell}[0],\dots,\mathbf{y}^{\mathrm{T}}_{\ell}[W-1]\right]^{\mathrm{T}}\in\mathbb{C}^{W N_R \times 1}$ collected from position $\bar{p}$:
\begin{equation}\label{eq:LR}
    \widehat{\mathbf{h}}_{LR,\ell} = \boldsymbol{\Pi}_L(\bar{p})\,\widehat{\mathbf{h}}_{LS,\ell},
\end{equation}
where $\widehat{\mathbf{h}}_{LR,\ell}\in\mathbb{C}^{W N_R N_T \times 1}$ is the LR-estimated channel vector and  $\widehat{\mathbf{h}}_{LS,\ell}=\mathbf{T}_{\ell} \,\mathbf{y}_{\ell}\in\mathbb{C}^{W N_R N_T \times 1}$ is the conventional LS MIMO channel estimate, whose analytical expressions are detailed in \cite{Cerutti2020}. Channel vector $\mathbf{h}$ (true or estimated) can be obtained from channel matrix $\boldsymbol{\mathcal{H}}$ (true or estimated) by vectorization $\mathbf{h}=\mathrm{vec}(\boldsymbol{\mathcal{H}})$.

The position-specific linear processing in \eqref{eq:LR} is designed in \cite{Nicoli2003} as:
\begin{equation}\label{eq:DST_projector}
    \boldsymbol{\Pi}_L(\bar{p}) = \mathbf{C}^{\frac{\mathrm{H}}{2}}\, \widehat{\boldsymbol{\Pi}}\, \mathbf{C}^{-\frac{\mathrm{H}}{2}},
\end{equation}
where
\begin{itemize}
    \item $\mathbf{C}\approx \frac{1}{\sigma^2_x} \,(\mathbf{I}_{W} \otimes \mathbf{I}_{N_T} \otimes \mathbf{Q}^{\mathrm{T}}_n)$ is the sample covariance matrix of the LS channel estimate $\widehat{\mathbf{h}}_{\mathrm{LS},\ell}$, needed to handle spatial/temporal noise correlations of interfering users in $\mathbf{Q}_n$;
    \item $\widehat{\boldsymbol{\Pi}}= \widehat{\mathbf{U}} \widehat{\mathbf{U}}^{\mathrm{H}}$ is the position-dependent projection matrix onto the ST propagation subspace associated to the ST basis (set of eigenmodes) 
    \begin{equation}\label{eq:compositeSSTbasis}
        \widehat{\mathbf{U}}=\widehat{\mathbf{U}}^*_{\mathrm{T}}\otimes \widehat{\mathbf{U}}^{\mathrm{Tx,*}}_{\mathrm{S}} \otimes \widehat{\mathbf{U}}^{\mathrm{Rx}}_{\mathrm{S}}.
    \end{equation} 
\end{itemize} 
Spatial ($\widehat{\mathbf{U}}^{\mathrm{Tx}}_{\mathrm{S}}\in\mathbb{C}^{N_T \times r^{\mathrm{Tx}}_{\mathrm{S}}}$, $\widehat{\mathbf{U}}^{\mathrm{Rx}}_{\mathrm{S}}\in\mathbb{C}^{N_R \times r^{\mathrm{Rx}}_{\mathrm{S}}}$) and temporal ($\widehat{\mathbf{U}}_{\mathrm{T}}\in\mathbb{C}^{W \times r_{\mathrm{T}}}$) MIMO channel eigenmodes are related to the set of DoDs, DoAs and delays, respectively. Eigenmodes form an orthonormal basis used to filter out from the LS estimate the noisy components that are not within the spanned algebraic subspace of the underlying channel. The eigenmodes $\widehat{\mathbf{U}}^{\mathrm{Tx}}_{\mathrm{S}}$, $\widehat{\mathbf{U}}^{\mathrm{Rx}}_{\mathrm{S}}$ and $\widehat{\mathbf{U}}_{\mathrm{T}}$ are estimated as the $r^{\mathrm{Tx}}_{\mathrm{S}}$, $r^{\mathrm{Rx}}_{\mathrm{S}}$ and $r_{\mathrm{T}}$ leading eigenvectors of the spatial (Tx and Rx) and temporal sample correlation matrices of the \textit{whitened} channel $\widetilde{\mathbf{h}}_{LS,\ell} = \mathbf{C}^{-\frac{\mathrm{H}}{2}} \widehat{\mathbf{h}}_{LS,\ell}$, computed over $L$ received pilot sequences from vehicular UEs passing on position $\bar{p}$. Notice that the directionality of the interference embedded in $\mathbf{Q}_n$ is typically estimated from the LS residual error \cite{Cerutti2020}. Therefore, matrix $\boldsymbol{\Pi}_L(\bar{p})$ operates a position-based, noise-aware modal filtering on the standard LS MIMO channel estimate.

LR performance is proportional to the \textit{sparsity degree} of the MIMO channel. It can be demonstrated that, if at least one of the following conditions holds \cite{mizmizi2021channel}:
\begin{align}\label{eq:}
  r^{\mathrm{Tx}}_{\mathrm{S}} < N_T, \;\;\;\;
  r^{\mathrm{Rx}}_{\mathrm{S}} < N_R, \;\;\;\;
  r_{\mathrm{T}} < W,
\end{align}
the LR method asymptotically ($L\rightarrow \infty$) outperforms LS. The value of $L$ for practical convergence depends on $N_T$, $N_R$ and $W$ as well as on the SNR. For the MIMO settings and bandwidths considered in Section \ref{sect:results}, $L\approx 100$ guarantees the convergence, that is for each location of the coverage cell. We remark that LR requires the knowledge of the UE position $\bar{p}$ during both the training phase (computation of $\boldsymbol{\Pi}_L(\bar{p})$) and the communication phase (run-time). The continuous exchange of position information in V2X systems is signalling intensive and increases the overhead on control channels. We explore in the following section a DL approach to retrieve the ST basis $\widehat{\mathbf{U}}$ (and $\boldsymbol{\Pi}_L(\bar{p})$) directly from $\widehat{\mathbf{h}}_{LS,\ell}$, without any explicit knowledge of the UE position.

\section{DL-based LR Channel Estimation}\label{sect:DL}

\begin{figure}[!t]
    \centering
    \includegraphics[width=\textwidth]{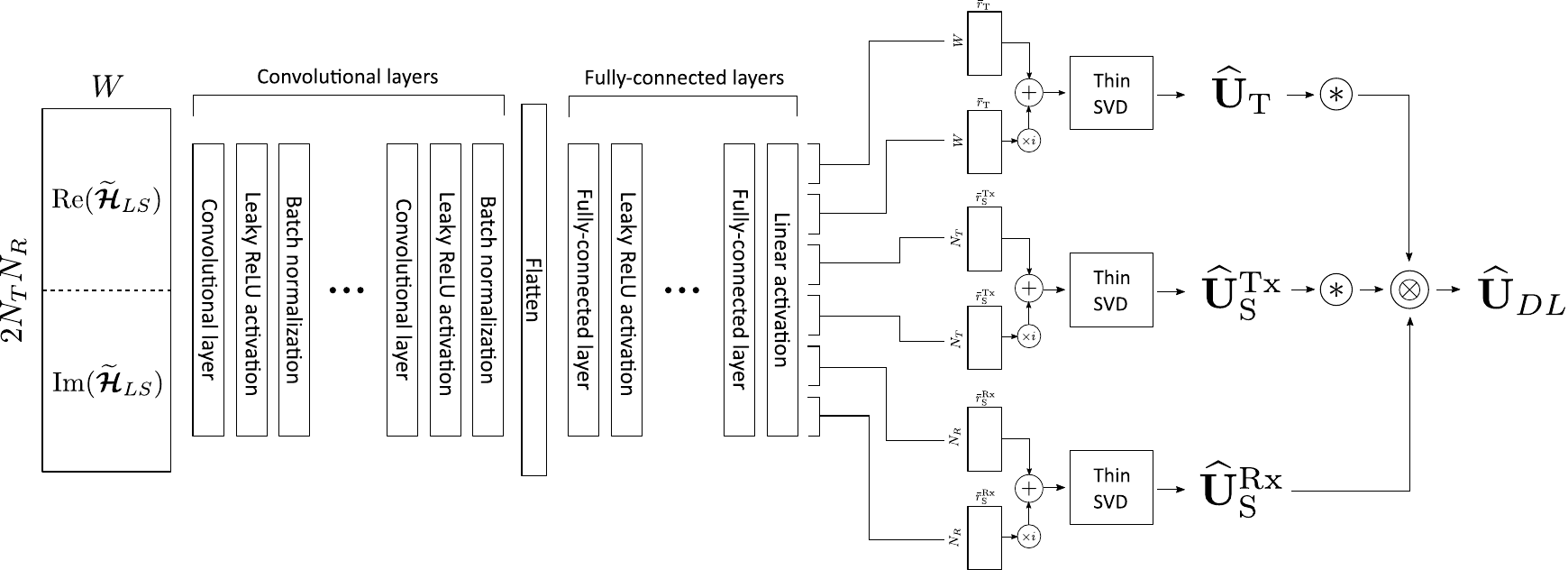}
    \caption{Proposed DNN architecture. The network takes as input the real and imaginary parts of a Least-Squares channel estimate, stacked on the spatial dimension, and outputs the corresponding Space-Time channel eigenmodes.}
    \label{fig:network_architecture}
\end{figure}

Leveraging the LR channel estimation algorithm described in Section \ref{sect:LR}, we propose a DNN to infer the spatial and temporal LR MIMO channel eigenmodes $\mathbf{U}_{\mathrm{T}}$, $ \mathbf{U}^{\mathrm{Tx}}_{\mathrm{S}}$, and $\mathbf{U}^{\mathrm{Rx}}_{\mathrm{S}}$ from a single received pilot sequence, or, equivalently, a LS MIMO channel estimate. 
Exploiting the representational power of DL, we test its capability to capture recurring vehicular patterns in the neighborhood of the BS within an urban scenario, without requiring the explicit signalling of UE's position.
In this regard, we use a large dataset of LS channel estimates $\{\widehat{\mathbf{h}}_{LS,m}\}_{m=1}^{m=M}$ gathered at the BS by multiple vehicular UEs along their path within the radio cell. The resulting labelled dataset $\{(\widehat{\mathbf{h}}_{LS,m},\, \widehat{\mathbf{h}}_{LR,m})\}_{m=1}^{m=M}$ is composed by couples associating a given input LS channel estimate $\widehat{\mathbf{h}}_{LS,m}$ to the corresponding LR channel estimate $\widehat{\mathbf{h}}_{LR,m}$, provided by the position-based LR method in Section \ref{sect:LR}. In this setting, $M$ denotes the cardinality of the dataset, comprising a suitable number of different tracks covering the whole radio cell. We assume the dataset is sufficiently large to apply the reference LR procedure and to train and evaluate the DNN.

The proposed DNN is depicted in Fig.~\ref{fig:network_architecture}. An input LS-estimated ST channel matrix $\widehat{\boldsymbol{\mathcal{H}}}_{LS}$ is first whitened as $\widetilde{\boldsymbol{\mathcal{H}}}_{LS} = \left(\mathbf{I}_{N_T} \otimes \frac{1}{\sigma^2_x}\mathbf{Q}^{-\frac{*}{2}}_n\right)\widehat{\boldsymbol{\mathcal{H}}}_{LS}$, then normalized by the maximum absolute value of its elements and finally stacked by real and imaginary parts along the spatial dimension, leading to a $2 N_T N_R \times W$ input matrix. We use a set of convolutional layers to extract effective features from the input channel matrix. Each convolutional layer employs the Leaky Rectified Linear Unit (Leaky ReLU) activation function \cite{maas2013rectifier}:
\begin{equation}
    \Gamma(x) = \begin{cases} 
        x & \text{for } x > 0\\
        0.01 x & \text{for } x \leq 0,
    \end{cases}
\end{equation}
and is followed by a batch normalization layer~\cite{ioffe2015batch}, which speeds up network convergence and improves stability. After flattening the output of the last batch normalization layer to a single vector of convolutional features, the latter are mapped through a set of fully-connected layers to six matrices representing (grouped in pairs) the real and imaginary parts of three complex-valued matrices with the same dimensions of $\widehat{\mathbf{U}}^{\mathrm{Tx}}_{\mathrm{S}}$, $\widehat{\mathbf{U}}^{\mathrm{Rx}}_{\mathrm{S}}$ and $\widehat{\mathbf{U}}_{\mathrm{T}}$ in \eqref{eq:compositeSSTbasis}, respectively. In order to output complex-valued unitary matrix representations, we project the three aforementioned complex-valued matrices on the corresponding Stiefel manifolds by applying thin Singular Value Decomposition (thSVD), which is an efficient operator to diagonalize LR matrices~\cite{golub2012matrix}. thSVD decomposes a matrix $\mathbf{A} \in \mathbb{C}^{n\times r}$, with $r \leq n$, as:
\begin{equation}\label{eq:SVD}
	[\mathbf{U}, \mathbf{s}, \mathbf{V}] = \text{thSVD}(\mathbf{A})\ \rightarrow\ \mathbf{A} = \mathbf{U}\,\text{diag}(\mathbf{s})\,\mathbf{V}^H,
\end{equation}
where $\mathbf{U} \in \mathbb{C}^{n \times r}$, $\mathbf{s} \in \mathbb{C}^{r\times 1}$, and $\mathbf{V} \in \mathbb{C}^{r \times r}$, with $\mathbf{U}$ and $\mathbf{V}$ unitary matrices. From \eqref{eq:SVD}, we consider only the $\mathbf{U}$ output, which is orthonormal and has the same dimensions as the target LR modes \eqref{eq:compositeSSTbasis}. Therefore, network training is carried only over $\mathbf{U}$, without updating the weights related to $\mathbf{s}$ and $\mathbf{V}$. Details on the automatic differentiation of complex-valued SVD can be found in \cite{wan2019automatic}. The DNN input-output relation is therefore described by the nonlinear parametric mapping
\begin{equation}
    \widehat{\mathbf{U}}_{DL} = f_{\boldsymbol{\Theta}}(\widetilde{\boldsymbol{\mathcal{H}}}_{LS}),
\end{equation}
where $\boldsymbol{\Theta}$ represents the network parameters to be optimized during training and $\widehat{\mathbf{U}}_{DL} = \widehat{\mathbf{U}}^*_{\mathrm{T}}\otimes \widehat{\mathbf{U}}^{\mathrm{Tx,*}}_{\mathrm{S}} \otimes \widehat{\mathbf{U}}^{\mathrm{Rx}}_{\mathrm{S}}$ is the DNN-inferred set of ST eigenmodes, aggregating the separate spatial and temporal eigenmodes as in \eqref{eq:compositeSSTbasis}. The LR-estimated MIMO channel is inferred as
\begin{equation}\label{eq:DNN_LR_prediction}
    \widehat{\mathbf{h}}_{LR}^{\mathrm{pred}} =\boldsymbol{\Pi}_{DL} \,\widehat{\mathbf{h}}_{LS}.
\end{equation}
where $\boldsymbol{\Pi}_{DL}=\mathbf{C}^{\frac{\mathrm{H}}{2}}\widehat{\mathbf{U}}_{DL} \widehat{\mathbf{U}}_{DL}^{\mathrm{H}} \mathbf{C}^{-\frac{\mathrm{H}}{2}}$ is the DL-estimated counterpart of the position-specific matrix $\boldsymbol{\Pi}_{L}(\bar{p})$ in \eqref{eq:LR}. Notice that $\boldsymbol{\Pi}_{DL}$ is not explicitly position-dependent.
The selected training loss function, to be minimized over the DNN parameters $\boldsymbol{\Theta}$, is the sum of the MSEs between the inferred LR channel estimates and the training ones:
\begin{equation}\label{eq:loss_function}
    \mathcal{L} = \sum_{m'=1}^{M'}\| \widehat{\mathbf{h}}_{LR,m'}^{\mathrm{train}} - \widehat{\mathbf{h}}_{LR,m'}^{\mathrm{pred}} \|^2,
\end{equation}
where $M'< M$ is the cardinality of the training dataset, a portion of the full one, $\widehat{\mathbf{h}}_{LR,m'}^{\mathrm{train}}$ is the $m'$-th point LR MIMO channel estimate used for training and $\widehat{\mathbf{h}}_{LR,m'}^{\mathrm{pred}}$ is from \eqref{eq:DNN_LR_prediction}.
In the simulations of Section \ref{sect:results}, the DNN parameters are optimized using the Adam \cite{kingma2014adam} optimizer, updating the network weights at mini-batches of 32 data points.

It is worth underlining that, differently from the MIMO channel eigenmodes obtained from the position-based LR method (Section \ref{sect:LR}), which have variable diversity orders in space, i.e., 
$\{r^{\mathrm{Tx}}_{\mathrm{S},m}, \,r^{\mathrm{Rx}}_{\mathrm{S},m},\, r_{\mathrm{T},m}\}_{m=1}^{m=M}$, all the unitary matrices inferred by the DNN have fixed orders $\bar{r}^{\mathrm{Tx}}_{\mathrm{S}},\,\bar{r}^{\mathrm{Rx}}_{\mathrm{S}},\,\bar{r}_{\mathrm{T}}$. Fixed orders are needed as the output layer of the DNN has fixed dimension, and this implies that each set of channel eigenmodes (spatial and temporal) lie on the same Stiefel manifolds \cite{absil2009optimization}. Notice that, considering a single cell scenario, the optimal value of $\bar{r}^{\mathrm{Tx}}_{\mathrm{S}},\,\bar{r}^{\mathrm{Rx}}_{\mathrm{S}},\,\bar{r}_{\mathrm{T}}$ should guarantee the best possible modal filtering provided by $\boldsymbol{\Pi}_{DL}$ over the \textit{whole} scenario. In principle, this shall imply to select the largest orders over the scenarios: $\bar{r}^{\mathrm{Tx}}_{\mathrm{S}}=\mathrm{max}\{r^{\mathrm{Tx}}_{\mathrm{S},m}\}_{m=1}^{m=M}$,  $\bar{r}^{\mathrm{Rx}}_{\mathrm{S}}=\mathrm{max}\{r^{\mathrm{Rx}}_{\mathrm{S},m}\}_{m=1}^{m=M}$, $\bar{r}_{\mathrm{T}}=\mathrm{max}\{r_{\mathrm{T},m}\}_{m=1}^{m=M}$. In practice, however, the true diversity orders of the channel are difficult to be estimated at each trajectory point within the cell. Moreover, the orders should be selected to enable a proper model transfer between different scenarios. Therefore, we consider the diversity orders as network hyperparameters to be optimized.

\section{Simulation Results}\label{sect:results}

\begin{figure*}
    \centering
    \subfloat{\includegraphics[width=0.23\textwidth]{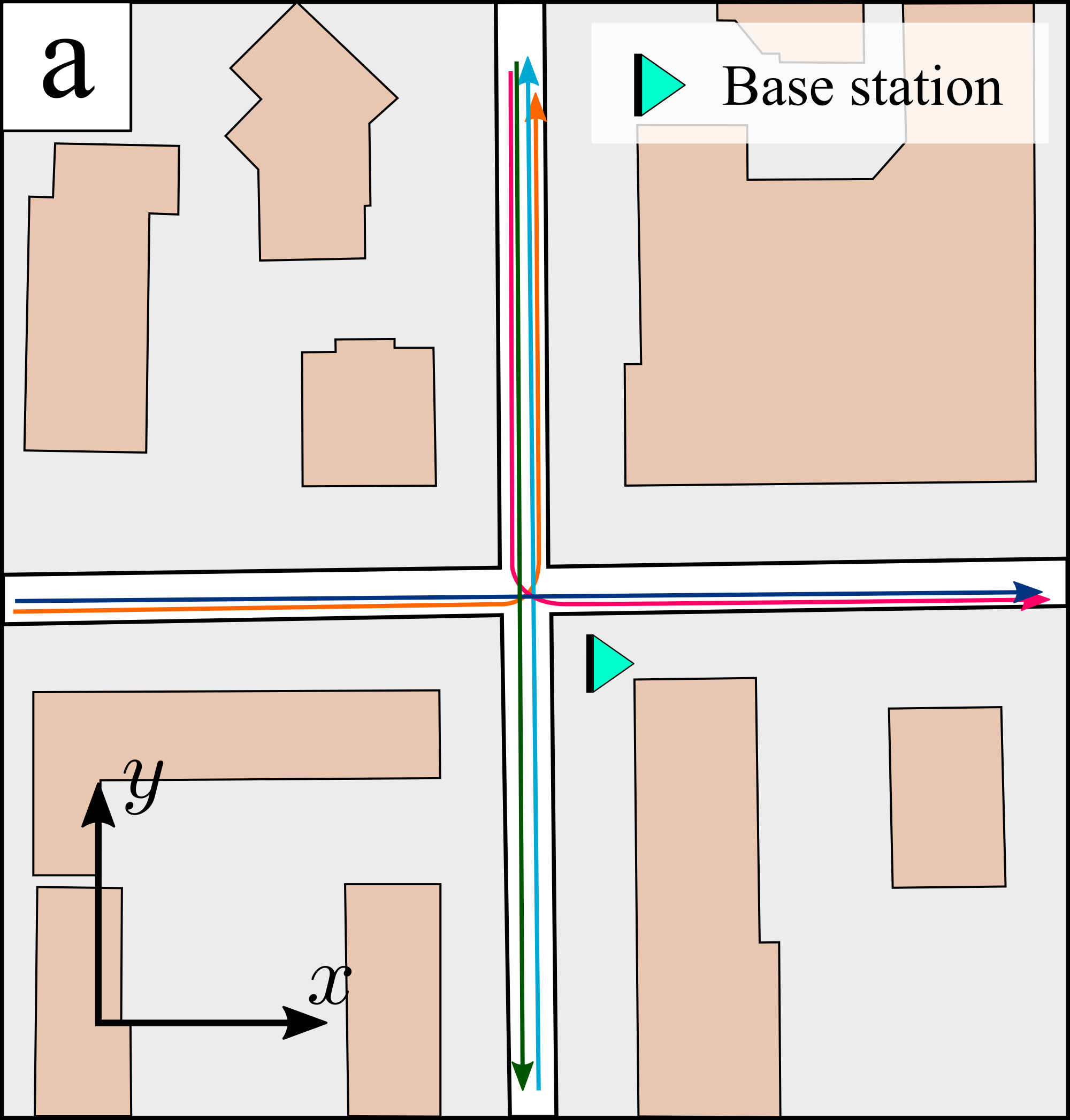}\label{fig:map_scenario_1}}
    \hspace{1cm}
    \subfloat{\includegraphics[width=0.23\textwidth]{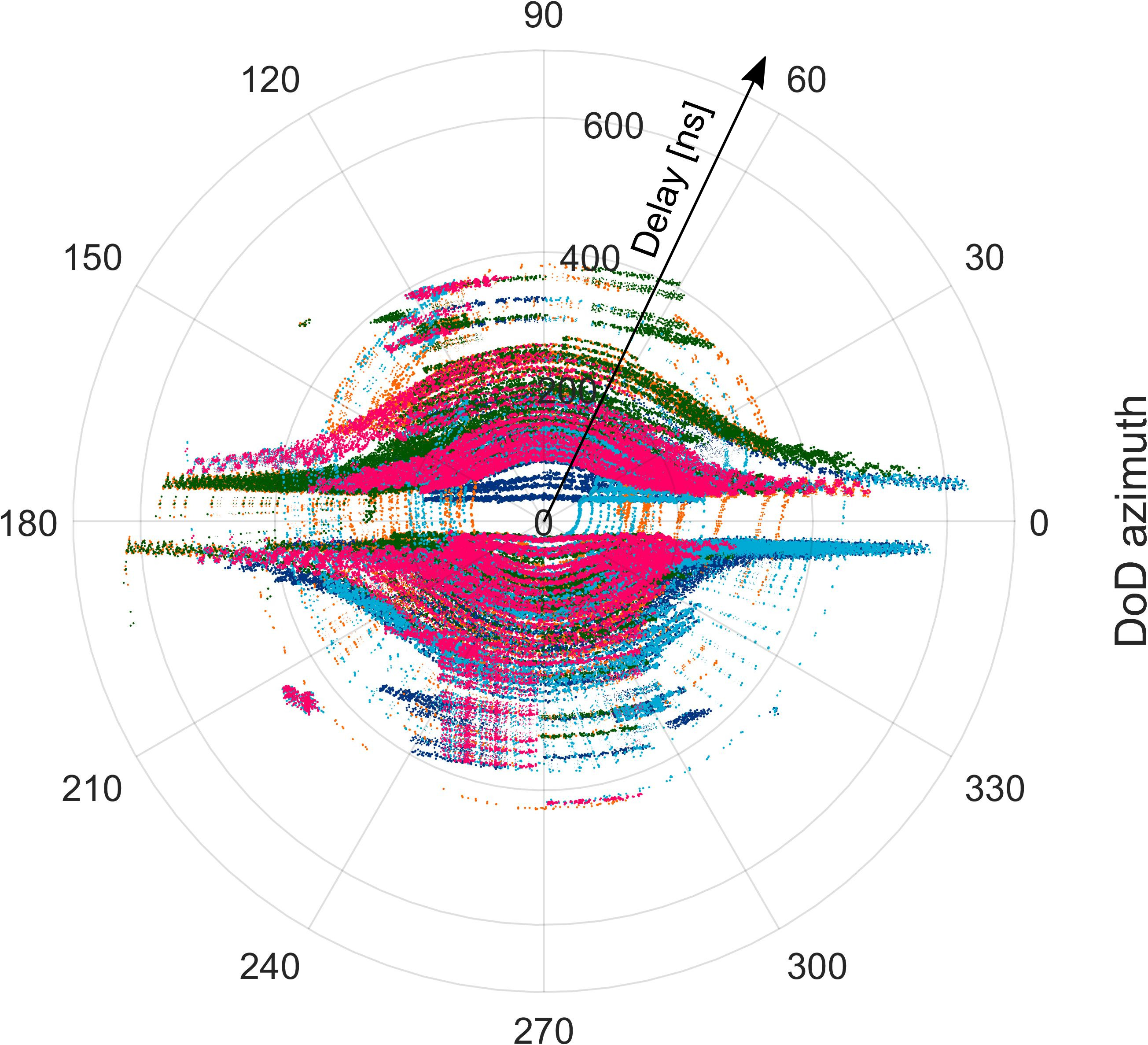}\label{fig:ST_params_scenario_1}}
    
    \subfloat{\includegraphics[width=0.23\textwidth]{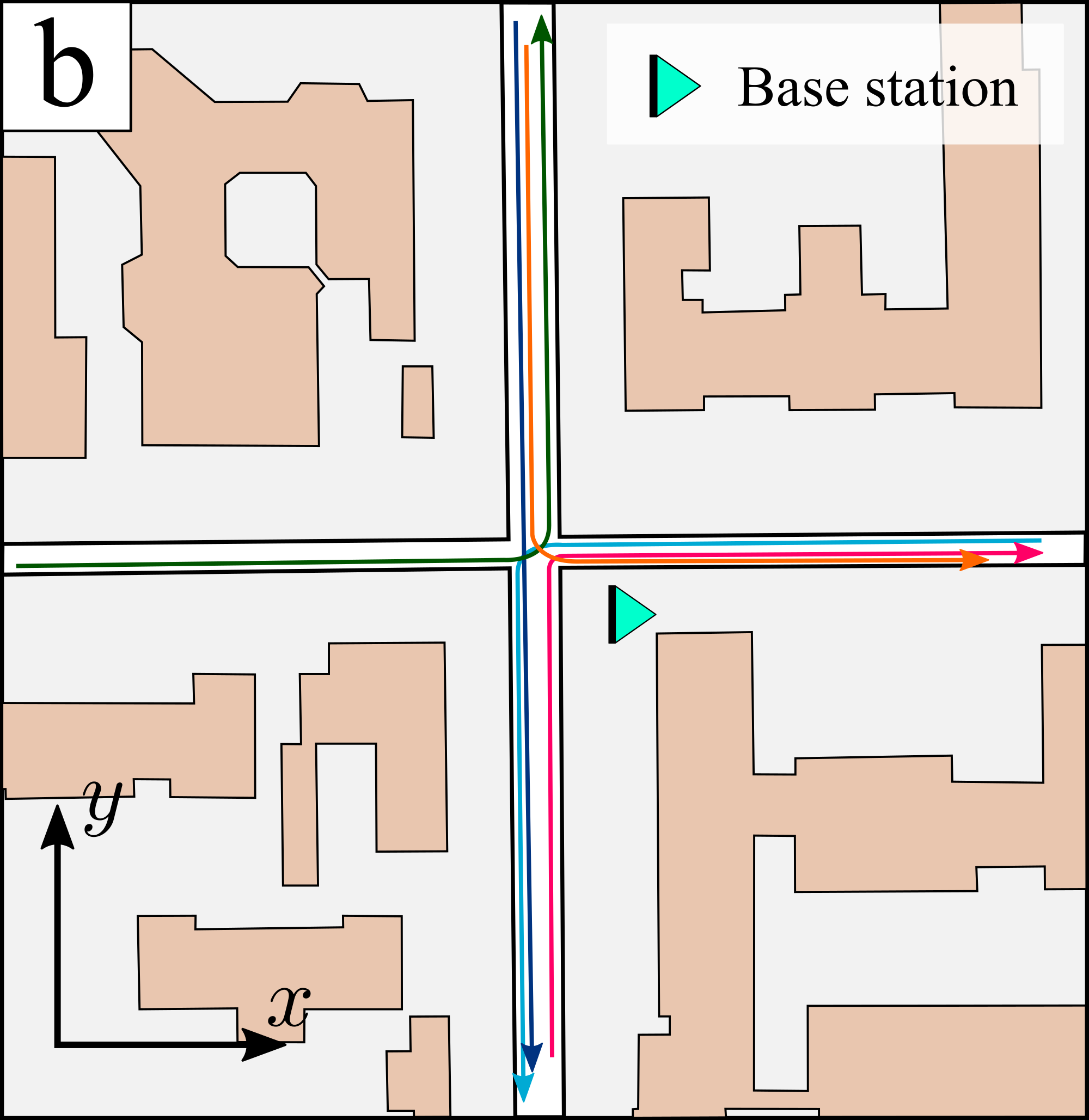}\label{fig:map_scenario_2}}
    \hspace{1cm}
    \subfloat{\includegraphics[width=0.23\textwidth]{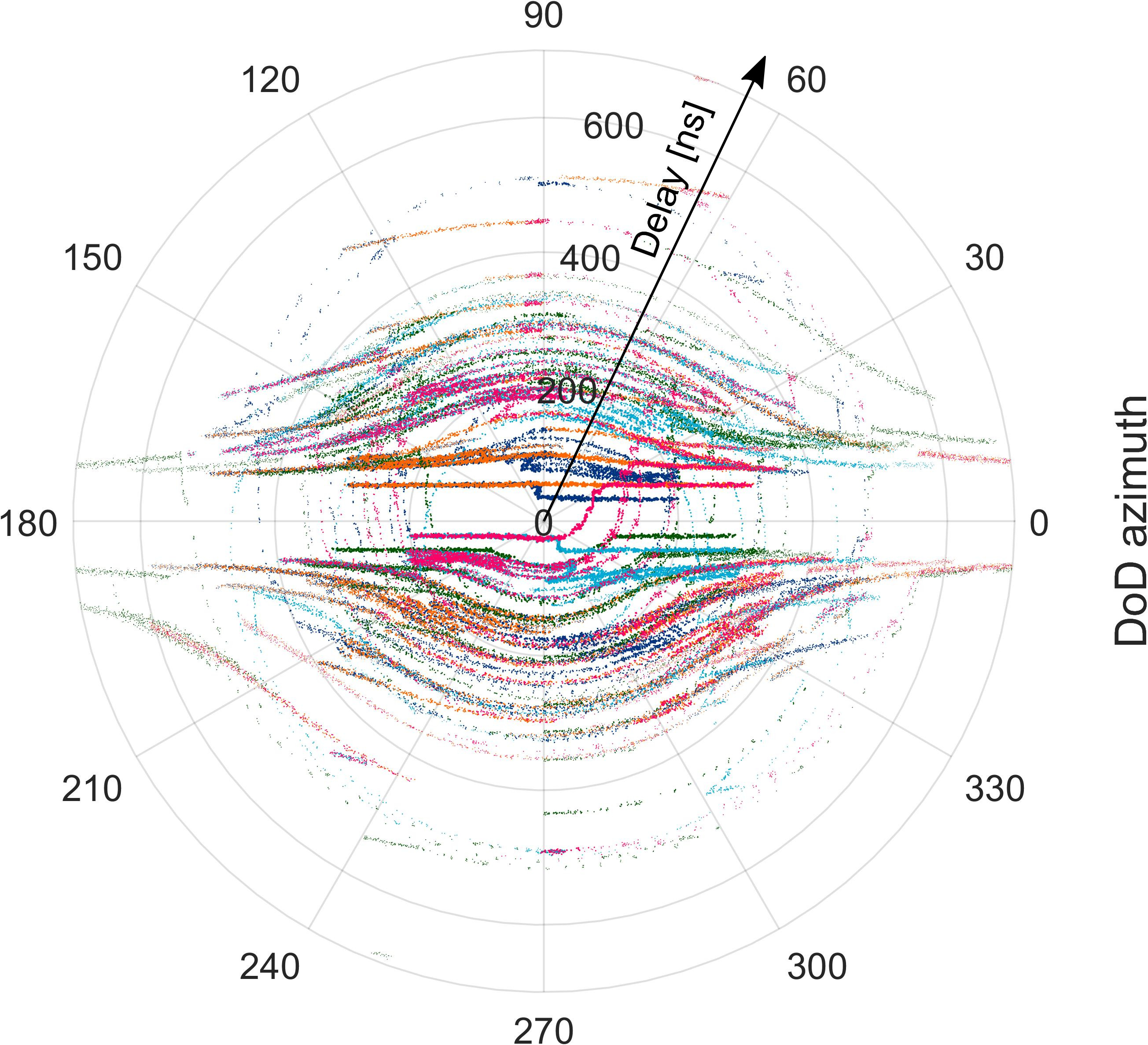}\label{fig:ST_params_scenario_2}}
    
    \subfloat{\includegraphics[width=0.23\textwidth]{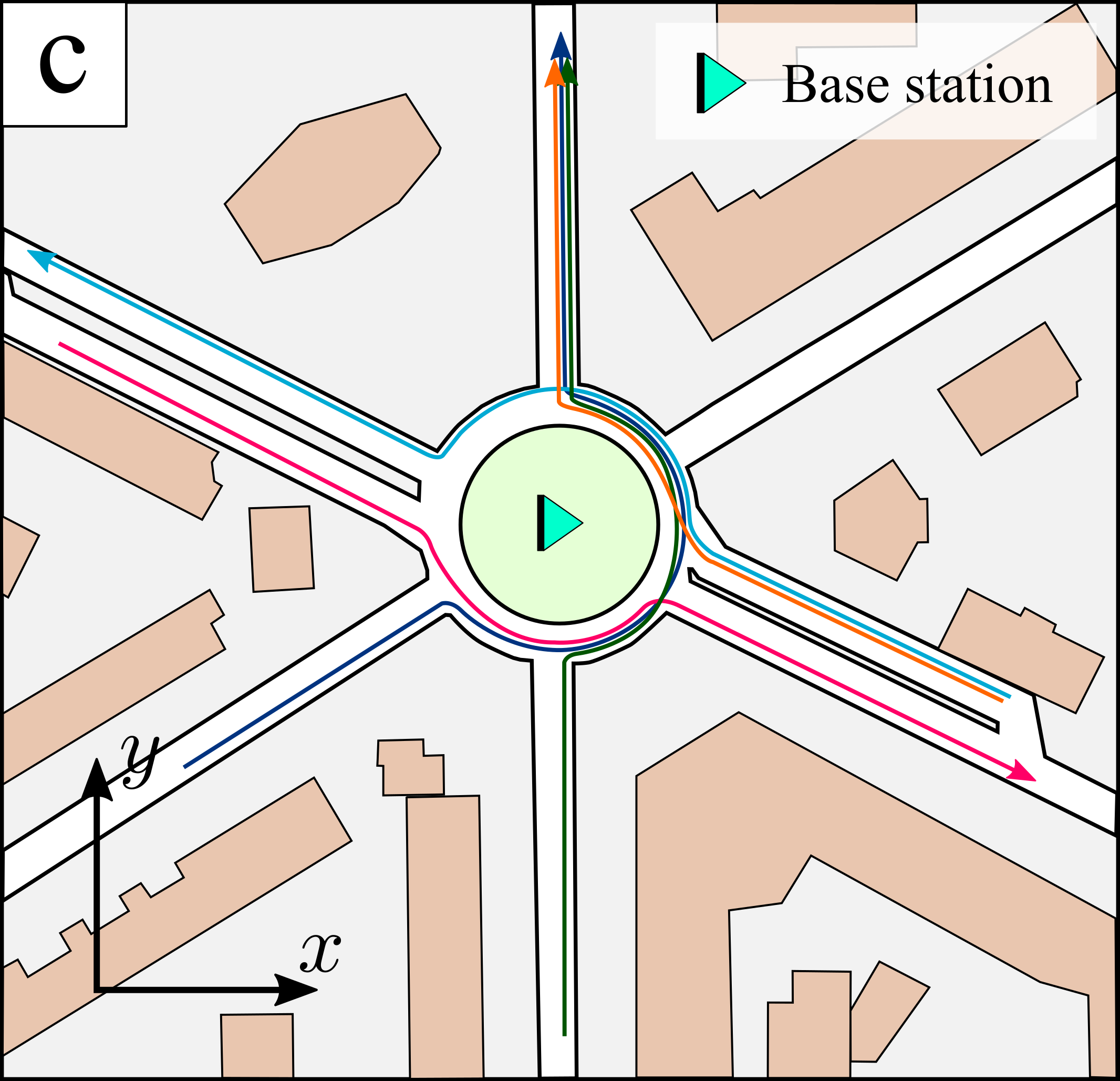}\label{fig:map_scenario_3}}
    \hspace{1cm}
    \subfloat{\includegraphics[width=0.23\textwidth]{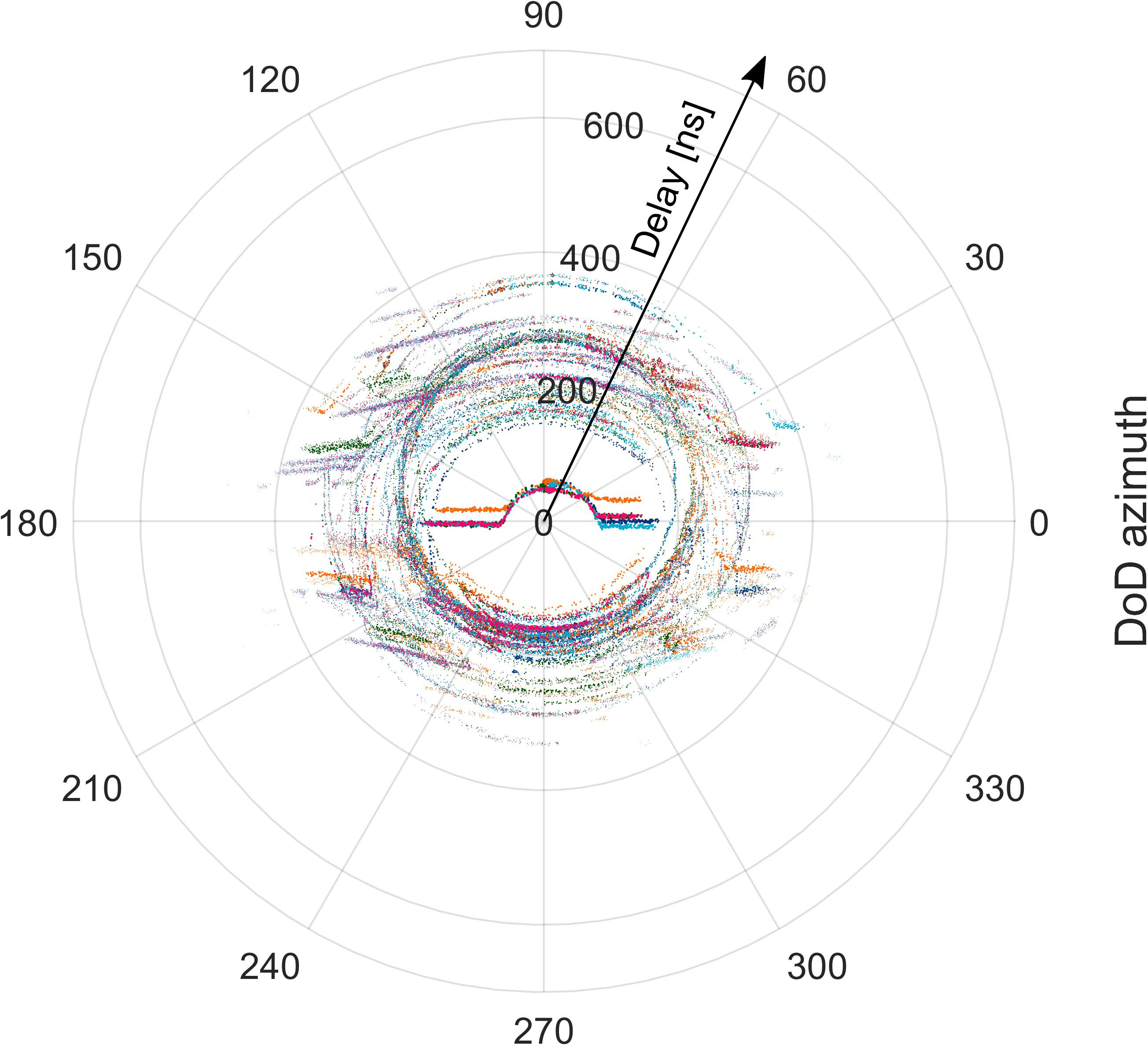}\label{fig:ST_params_scenario_3}}
    
    \subfloat{\includegraphics[width=0.23\textwidth]{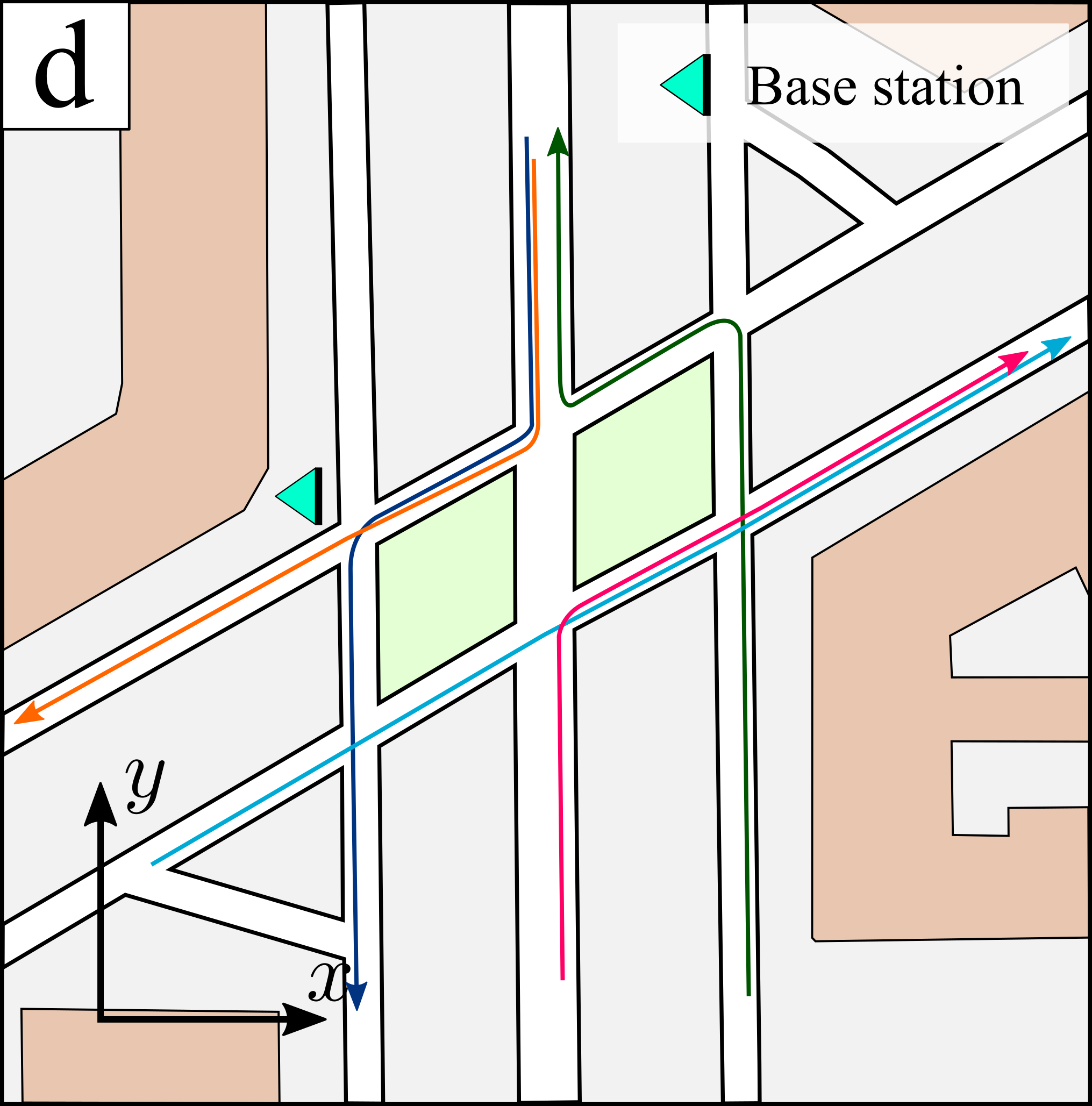}\label{fig:map_scenario_4}}
    \hspace{1cm}
    \subfloat{\includegraphics[width=0.23\textwidth]{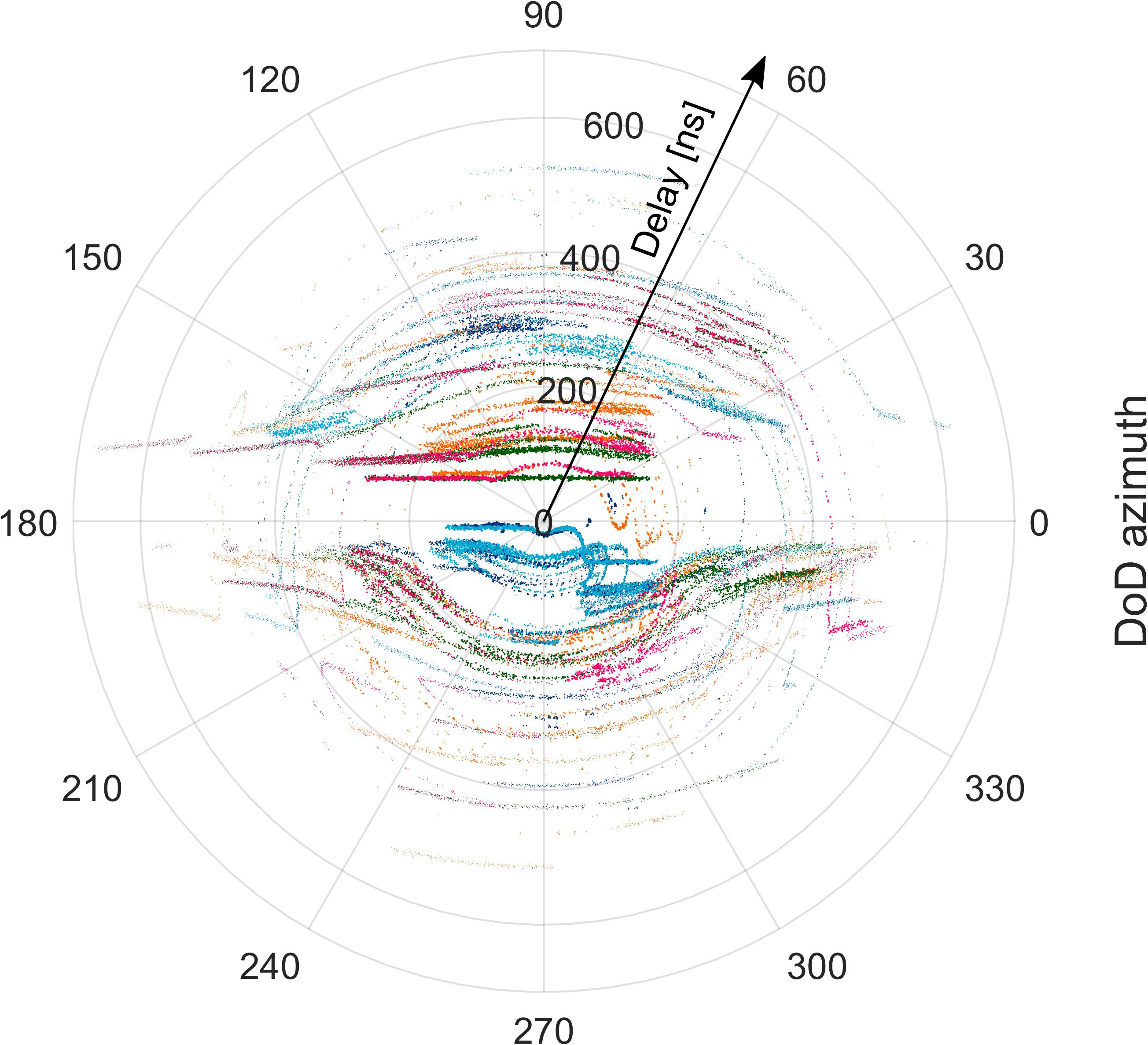}\label{fig:ST_params_scenario_4}}
    
    \subfloat{\includegraphics[width=0.23\textwidth]{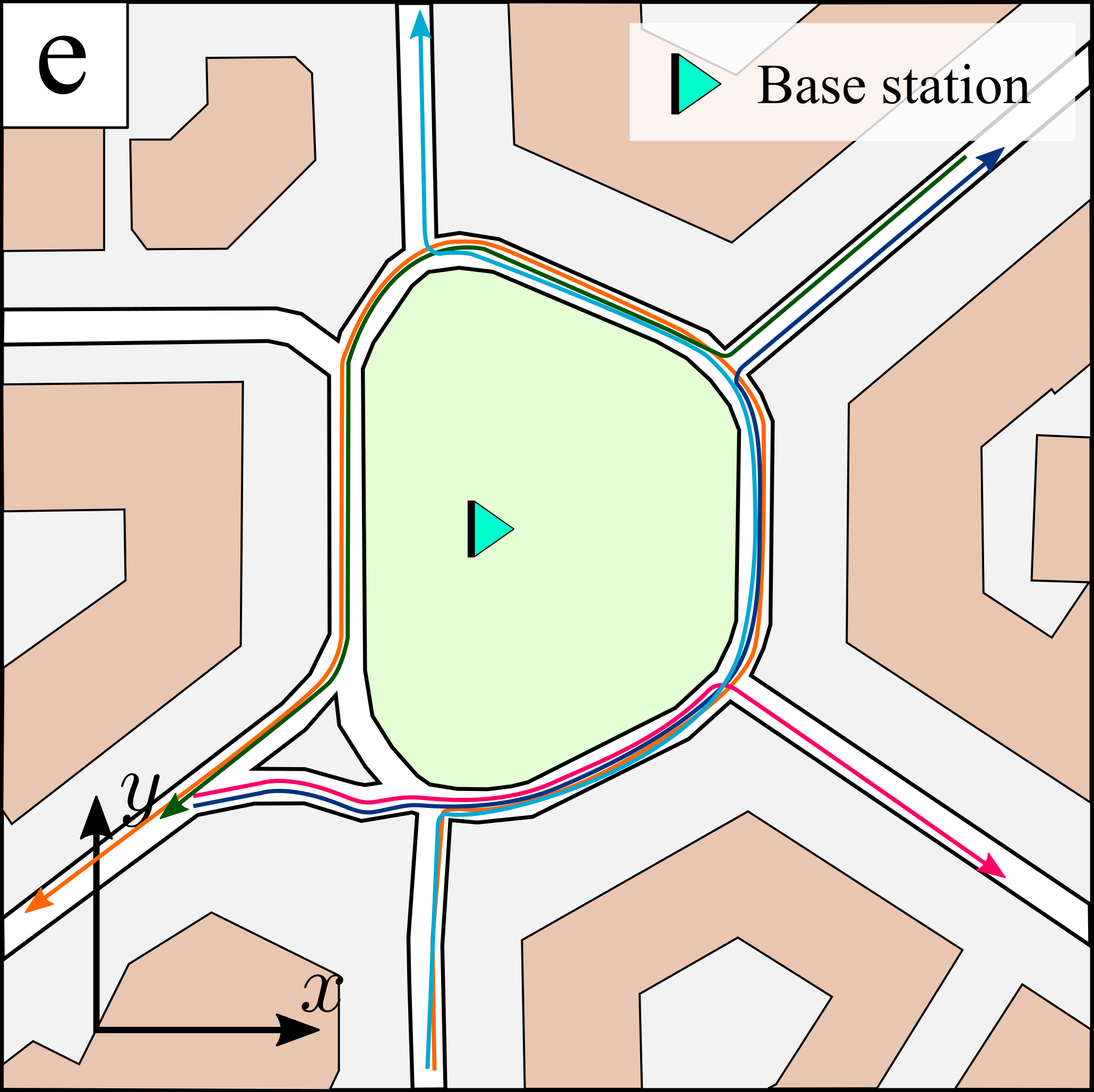}\label{fig:map_scenario_5}}
    \hspace{1cm}
    \subfloat{\includegraphics[width=0.23\textwidth]{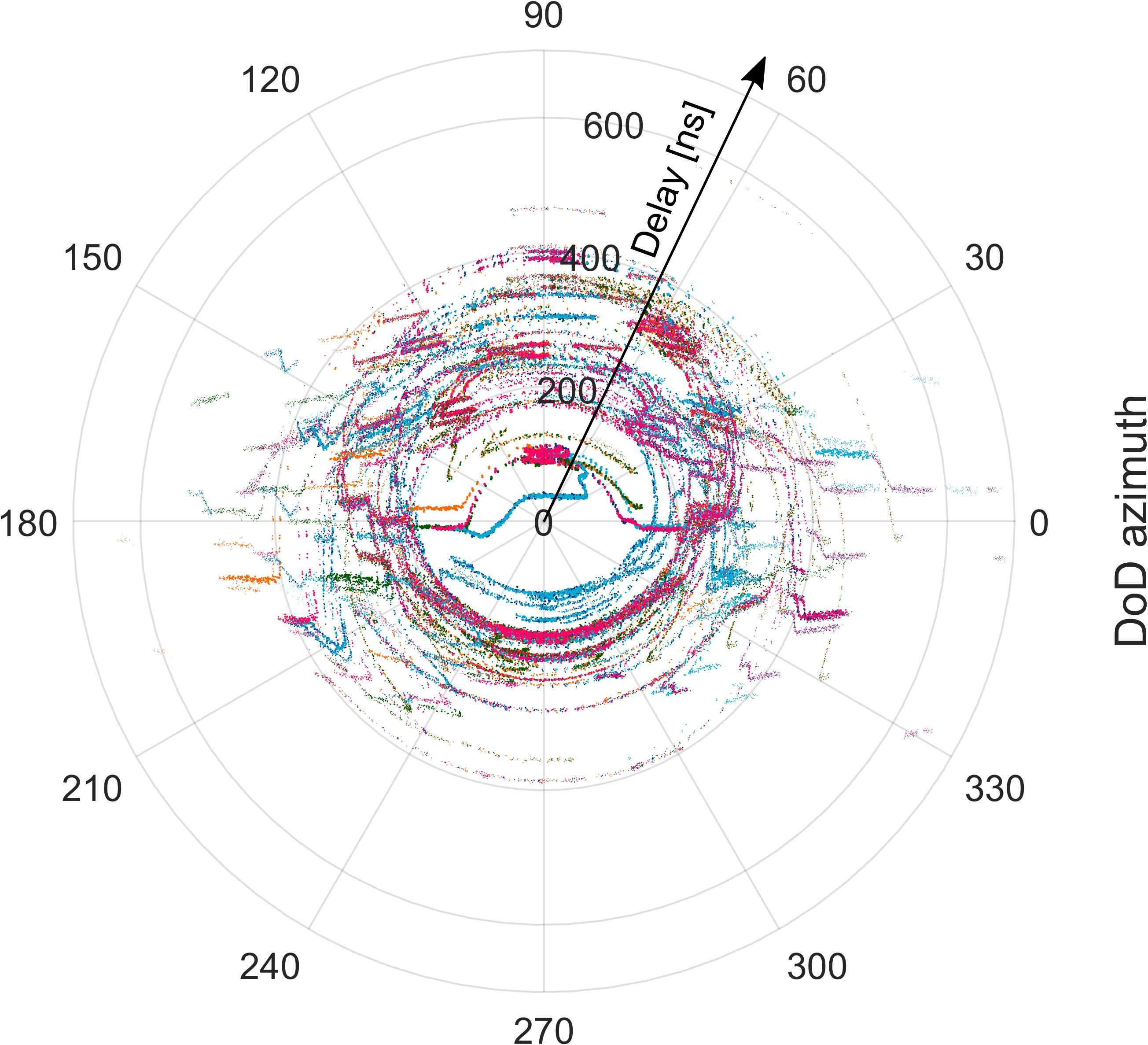}\label{fig:ST_params_scenario_5}}
    
    \caption{Scenarios for the training and evaluation of the proposed DNN, with corresponding DoD/delay channel features (ray-tracing derived). Colored lines represent the reference vehicle trajectories, associated to colored paths in the DoD/delay plots.}
    \label{fig:scenarios_for_TL}
\end{figure*}

In this section, we present numerical results proving the effectiveness of the proposed DL-based LR channel estimation. Five scenarios are selected for numerical testing from portions of the city of Milan. They are depicted in Fig. \ref{fig:scenarios_for_TL}, representing typical urban road crossings characterized by LOS propagation. Each scenario has a different geometry, road topology, and vehicular trajectory patterns, thus leading to diverse features in the ST domain. 
The simulation parameters are presented in Table \ref{tab:simulation_parameters}. We consider a OFDMA uplink communication at $f_0=28$ GHz carrier frequency between multiple vehicular UEs and a tri-sectoral BS, the former equipped with $N_T = 16$ ($4 \times 4$) antennas and the latter with $N_R=64$ ($8 \times 8$) antennas (for each sector). The BS is located at $6$ m from ground, in the position highlighted with a triangle in Fig. \ref{fig:scenarios_for_TL}, while each UE moves at $1.5$ m from ground. Two different communication bandwidths per UE are tested: $B=1$ MHz, for which the channel is frequency-flat ($W = 1$), and $50$ MHz, producing a frequency-selective channel ($W \gg 1$) in each of the five scenarios. We analyze the performance of the proposed channel estimation method in terms of Normalized Mean Squared Error (NMSE), defined as
\begin{equation}\label{eq:MSE}
    \mathrm{NMSE} = \frac{\mathbb{E}[\lVert\mathbf{h}_{\ell} - \widehat{\mathbf{h}}_{LR,\ell}\rVert^2]}{\mathbb{E}[\lVert\mathbf{h}_{\ell} - \widehat{\mathbf{h}}_{LS,\ell}\rVert^2]},
\end{equation}
to highlight the MSE gain of LR compared to LS as reference method.

\begin{table}[t!]
\centering
\caption{Simulation parameters}
\begin{tabular}{l c c}
\toprule
\textbf{Parameter} & \textbf{Symbol} & \textbf{Value}\\
\noalign{\smallskip}
\hline
\noalign{\smallskip}
Carrier frequency & $f_0$ & $28$ GHz\\\noalign{\smallskip}
Bandwidth & $B$ & $1$, $50$ MHz\\\noalign{\smallskip}
BS height from the ground & - & $6$ m\\\noalign{\smallskip}
UEs height from the ground & - & $1.5$ m\\\noalign{\smallskip}
Number of BS antennas & $N_R$ & $64$ ($8 \times 8$) \\\noalign{\smallskip}
Number of UE antennas & $N_T$ & $16$ ($4 \times 4$)\\\noalign{\smallskip}
Signal to Noise Ratio & SNR & $0$ dB \\\noalign{\smallskip}
\bottomrule
\end{tabular}
\label{tab:simulation_parameters}
\end{table}

\subsection{Simulation setup}

The datasets used for training the DNN over each scenario are produced by means of simulated channel data over realistic vehicle trajectories, obtained from  SUMO (Simulation of Urban MObility) \cite{SUMO2018}. The mmWave channel parameters at $28$ GHz are simulated by ray tracing using Altair WinProp \cite{altairwinprop} software, which provides for each considered geographical point the Direction of Departure (DoD) $\psi$, the Direction of Arrival (DoA) $\theta$, the power $\Omega$ and the scattering amplitude $\alpha$ of each ray. The MIMO channel impulse response follows from \eqref{eq:channel_matrix_tap_time} by fixing the maximum number of taps over all the five scenarios ($W=22$ for $B=50$ MHz, determined by ray tracing). This enables the direct model transfer from one scenario to another.  
For each scenario, a dataset of $M=2.5 \times 10^5$ channel samples has been produced considering multiple (different) realizations of 5 reference vehicular trajectories, where the Signal-to-Noise Ratio (SNR) has been fixed at $0$ dB along all the trajectory. 

We train the proposed model on scenario a), testing the learning capabilities of the DNN by comparing the NMSE of the DL-based LR method against the NMSE of the reference position-based LR introduced in Section \ref{sect:LR}. Then, we analyse the generalization of the model to the remaining 4 scenarios b), c), d) and e) considering two distinct procedures: \textit{(i)} testing the performance of the trained model by directly applying it to the new 4 scenarios, \textit{without any retraining}; \textit{(ii)} fine-tuning of the model trained on a) on the specific application scenario (b,c,d,e) by training only the last two fully-connected layers of the DNN (only for $B=50$ MHz, since no improvement has been obtained by applying this procedure to the frequency-flat case). The DNN is trained using Adam optimizer  \cite{kingma2014adam} with a learning rate $\eta = 0.001$.

\subsection{Results for $B=1$ MHz (frequency-flat)}
 
We present the results obtained by applying the proposed DL-based channel estimation to frequency-flat MIMO channels with $W = 1$ temporal tap. After showing the NMSE performance of the model on the reference scenario a) (Fig. \ref{fig:scenarios_for_TL}), we examine its generalization capabilities by directly applying it over scenarios b), c), d), e).

\subsubsection{Performance of the DNN model on the reference urban scenario}

\begin{figure}[!t]
    \centering
    \subfloat{\includegraphics[width=0.31\textwidth,valign=t]{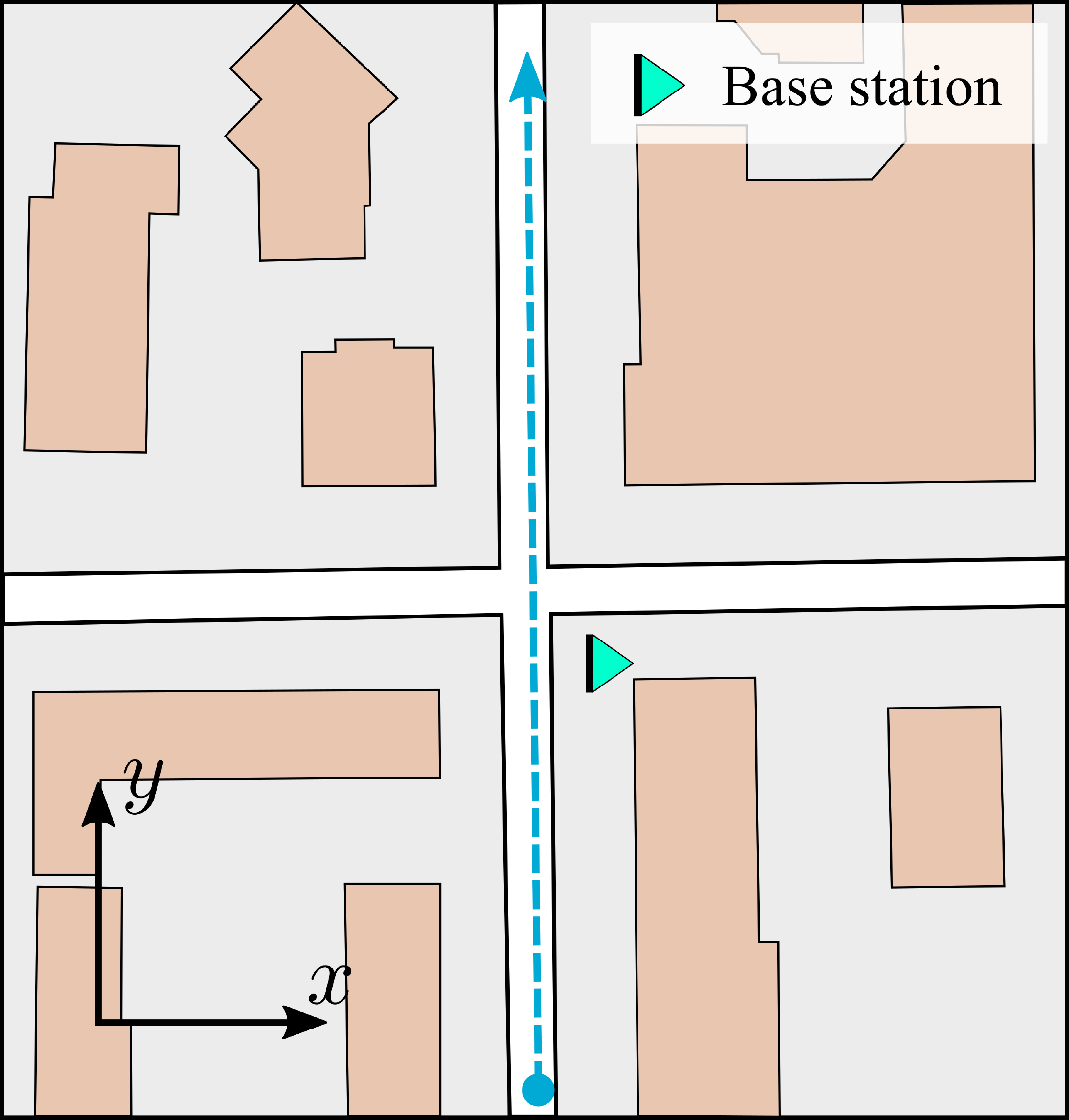}}
    \hspace{.5cm}
    \subfloat{\includegraphics[width=0.45\textwidth,valign=t]{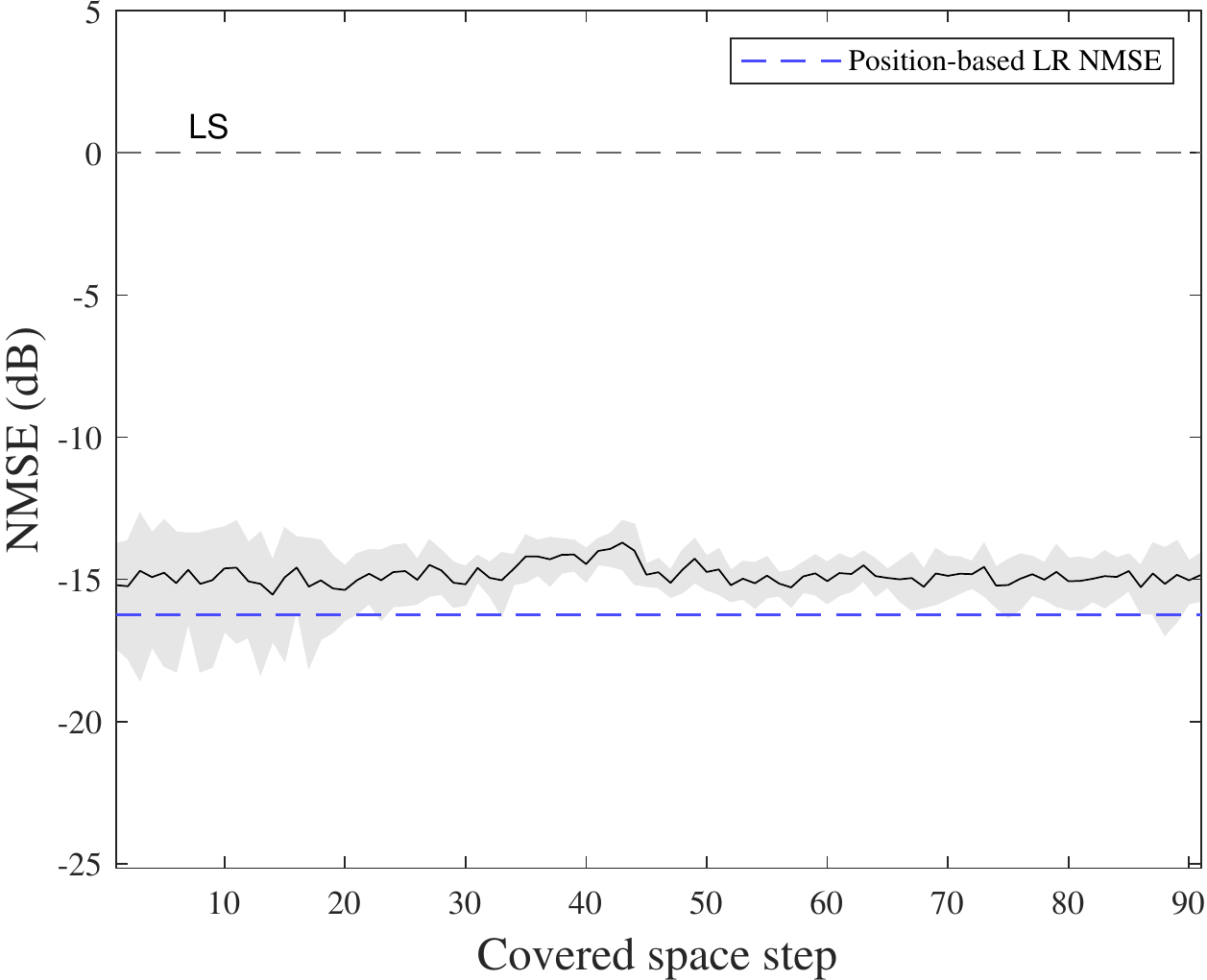}}\\
    \vspace{.5cm}
    \subfloat{\includegraphics[width=0.31\textwidth,valign=t]{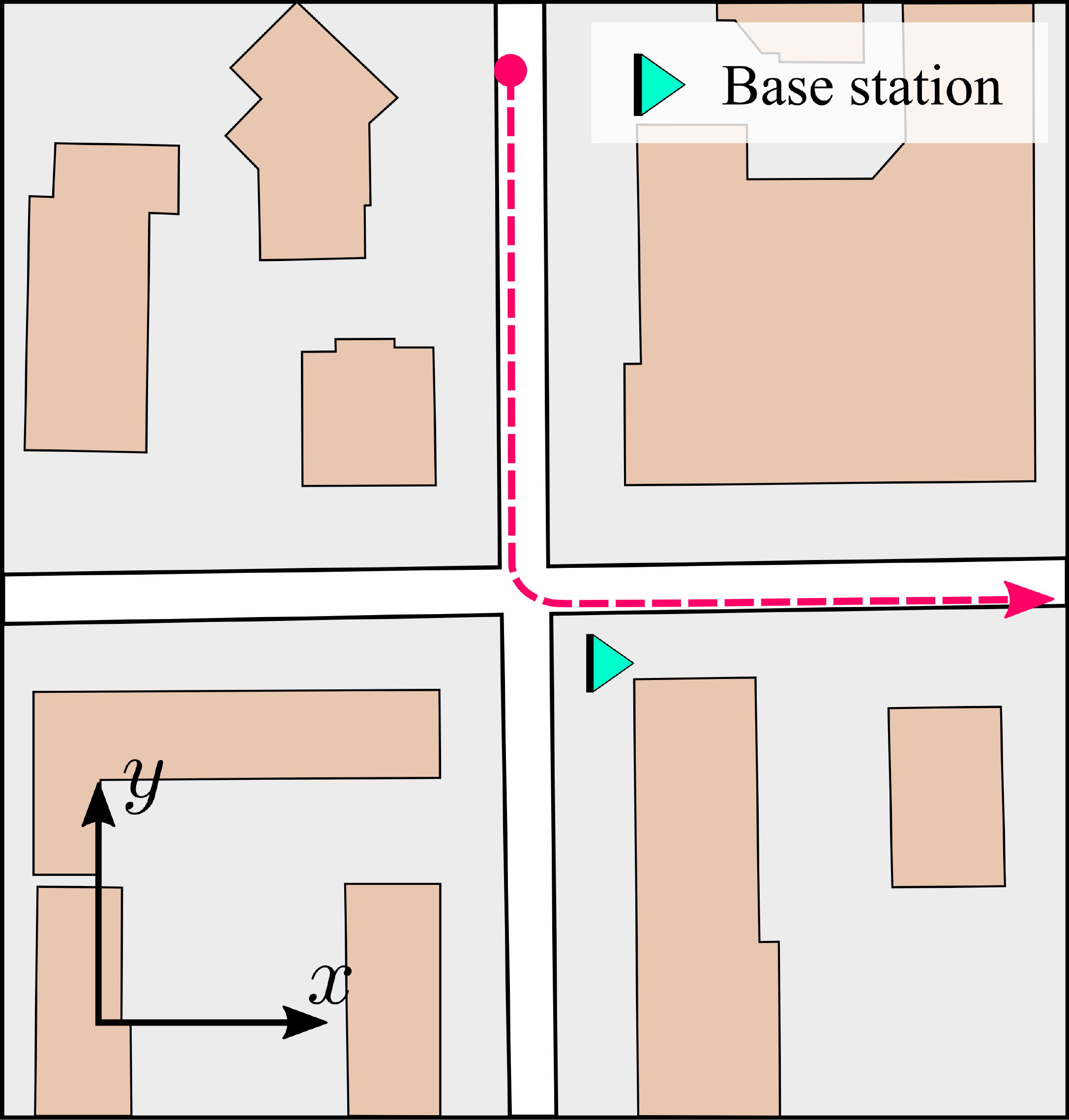}}
    \hspace{.5cm}
    \subfloat{\includegraphics[width=0.45\textwidth,valign=t]{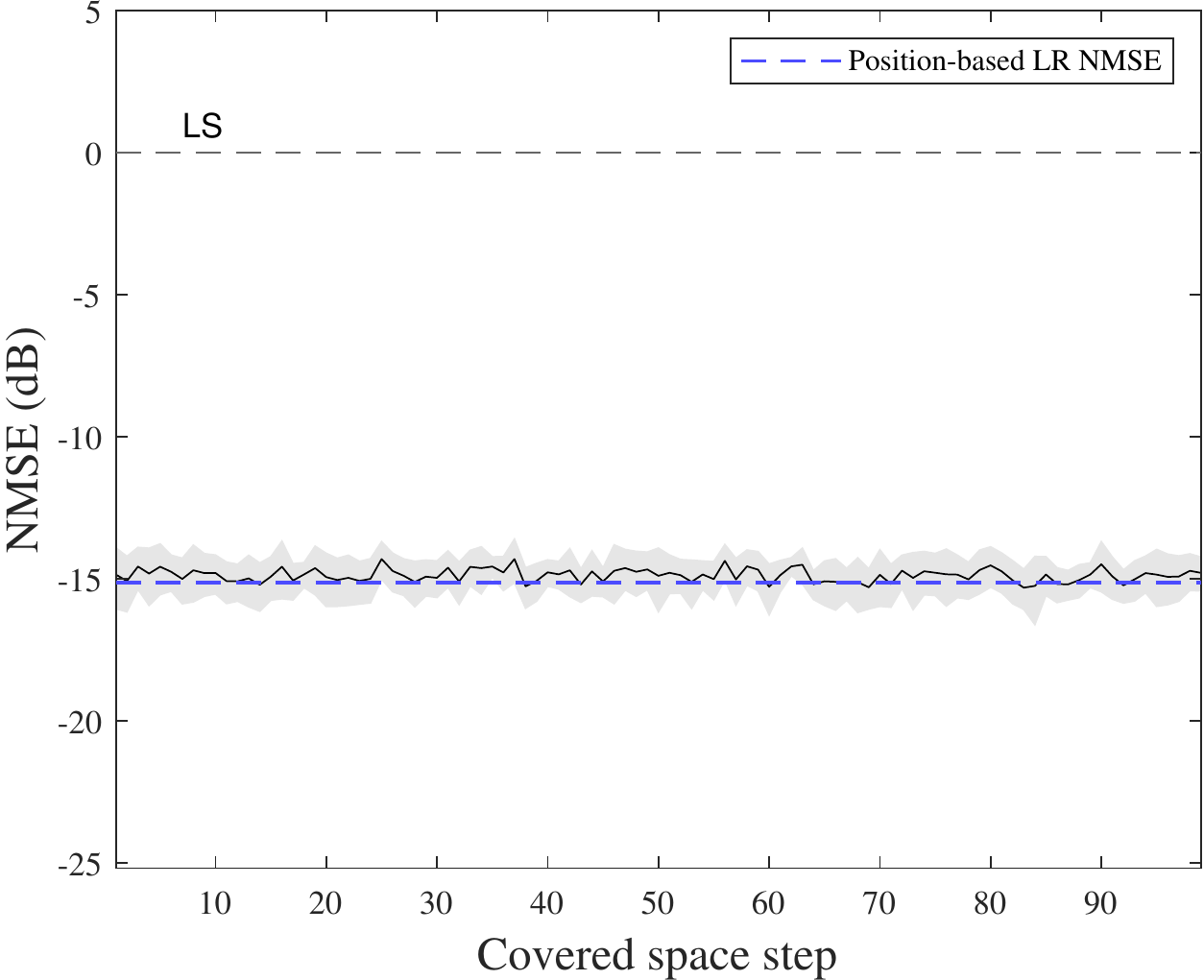}}
    
    \caption{Evaluation of the proposed DL-based channel estimator over sample trajectories in the reference scenario a) for $B=1$ MHz.}
    \label{fig:results_over_sample_trajectories_1MHz}
\end{figure}

The DNN model selected for frequency-flat channel estimation has 3 convolutional layers and 4 fully-connected layers. The first two convolutional layers use $64$ filters while the last one uses a single filter. All the three convolutional layers use a $1 \times 1$ kernel, while the fully-connected layers are composed of $50$ neurons each. This leads to $\approx 2.5 \times 10^5$ trainable model parameters $\boldsymbol{\Theta}$. We train our model with LS and LR channel estimates gathered on the reference scenario depicted in Fig. \ref{fig:scenarios_for_TL}a). By hyperparameter search, we selected the ranks $\bar{r}^{\mathrm{Tx}}_{\mathrm{S}} = 4$,  $\bar{r}^{\mathrm{Rx}}_{\mathrm{S}} = 8$, $\bar{r}_{\mathrm{T}} = 1$ for the inferred unitary matrices corresponding to the spatial and temporal MIMO channel eigenmodes. Hence, with $\bar{r}_{\mathrm{T}} = 1$, the MIMO channel is characterized by spatial modes only.
The DL model converges within 10 training iterations to an average NMSE value of $-14.9$ dB (MSE gain with respect to LS estimation), to be compared with $-15.7$ dB NMSE provided by the position-based LR method in Section \ref{sect:LR}. Fig. \ref{fig:results_over_sample_trajectories_1MHz} shows the NMSE performance of the proposed DL model when applied to 2 reference vehicular trajectories within the training scenario a).
We consider multiple realizations of each trajectory to estimate pointwise the NMSE standard deviation for the inferred LR channel estimates (represented by the shaded gray area in Fig. \ref{fig:results_over_sample_trajectories_1MHz}). The blue dashed line is instead the mean NMSE provided by position-based LR channel estimation described in Section \ref{sect:LR} (used for training), averaged over the whole length of the chosen trajectory. The results show that the DL-based NMSE closely matches the position-based NMSE except for some small performance penalty ($< 2$ dB). The same behavior has been also observed on the other 3 trajectory types over which the DNN model has been trained.

\subsubsection{Generalization of the DNN model to different urban scenarios}

To assess the effectiveness of the proposed DL method when challenged with new ST features of the environment, we test the model trained on the reference scenario a) against the b), c), d), and e) environments in Fig. \ref{fig:scenarios_for_TL}. Notice that no transfer learning fine-tuning is used here. Our aim is to evaluate the capability of the model to map local convolutional features---learned from channel impulse responses sampled on the reference scenario---to the spatial and temporal MIMO channel eigenmodes on new data.
\begin{figure}[!t]
    \centering
    \includegraphics[width=0.5\textwidth]{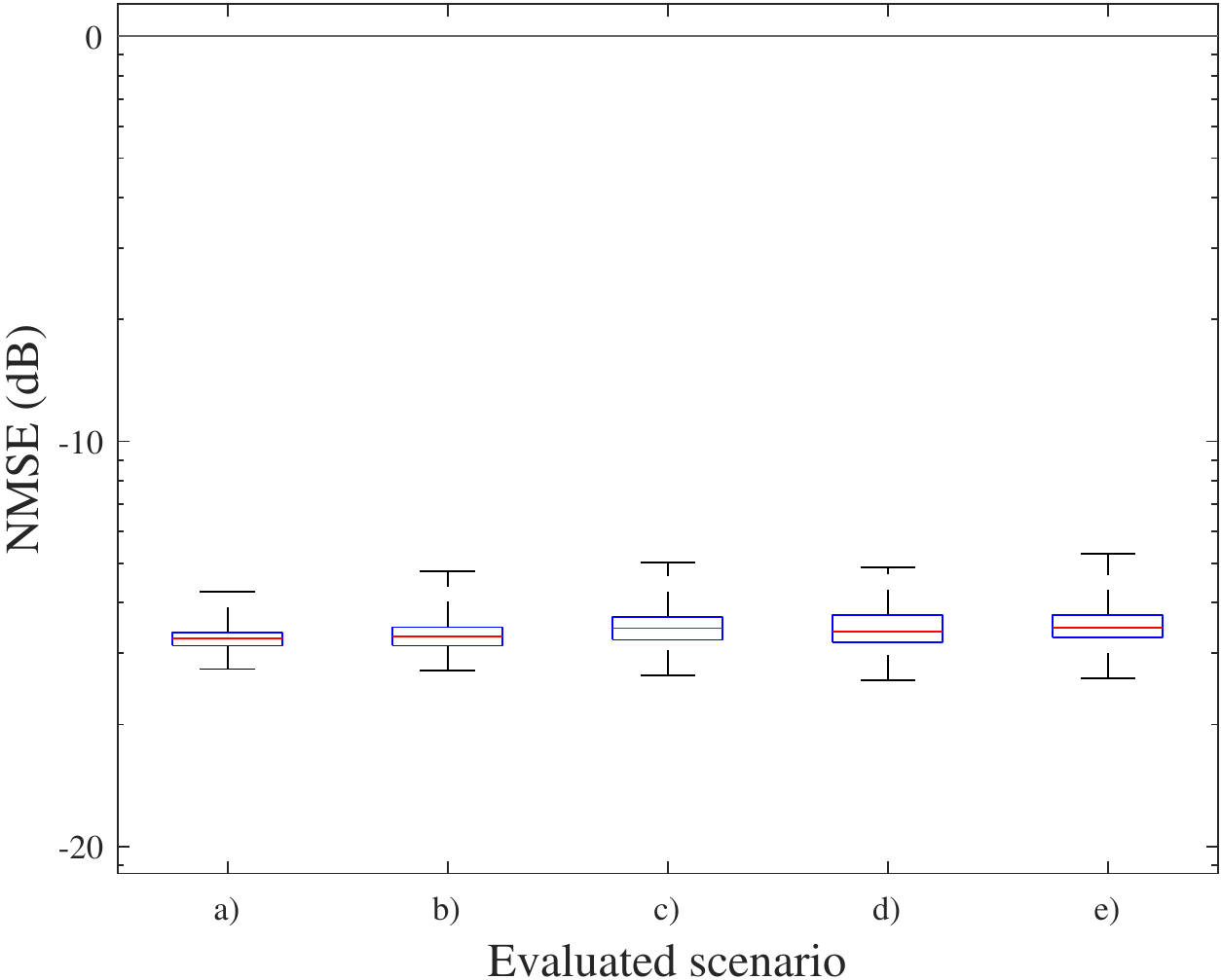}
    \caption{Box plots showing median, interquartile range, and total range for the NMSE achieved using the DNN trained on scenario a) over the remaining scenarios for $B=1$ MHz.}
    \label{fig:box_plots_1MHz}
\end{figure}
Fig. \ref{fig:box_plots_1MHz} summarizes the NMSE of the channel estimates inferred over the tested scenarios by means of box plots, where the red line represents the median, the box encloses the interval between the first and the third quartiles, and the outer bars delimit the range of observed NMSE performances. We notice that the DL model transfer between one scenario to the others provides comparable NMSE performance, with only a slight increase of the NMSE dispersion. We also observed that TL fine-tuning does not provide any benefit, as the DL model is able to represent the MIMO spatial eigenmodes with the same accuracy experienced on a reference scenario. This result is particularly relevant for the implementation of the proposed DL-based channel estimation in practical systems, as it allows a remarkable reduction of the number of collaborative vehicles (UEs) used for training the DNN, at least for the frequency-flat channel case. In the considered settings, the DNN training dataset can be reduced by $\approx 80$ \%, as a full re-training of the DNN over the other 4 scenarios is not necessary.

\subsection{Results for $B=50$ MHz (frequency-selective)}

\begin{figure}
    \centering
    \subfloat{\includegraphics[width=0.31\textwidth,valign=t]{images/results/maps/map_1.pdf}}
    \hspace{.5cm}
    \subfloat{\includegraphics[width=0.45\textwidth,valign=t]{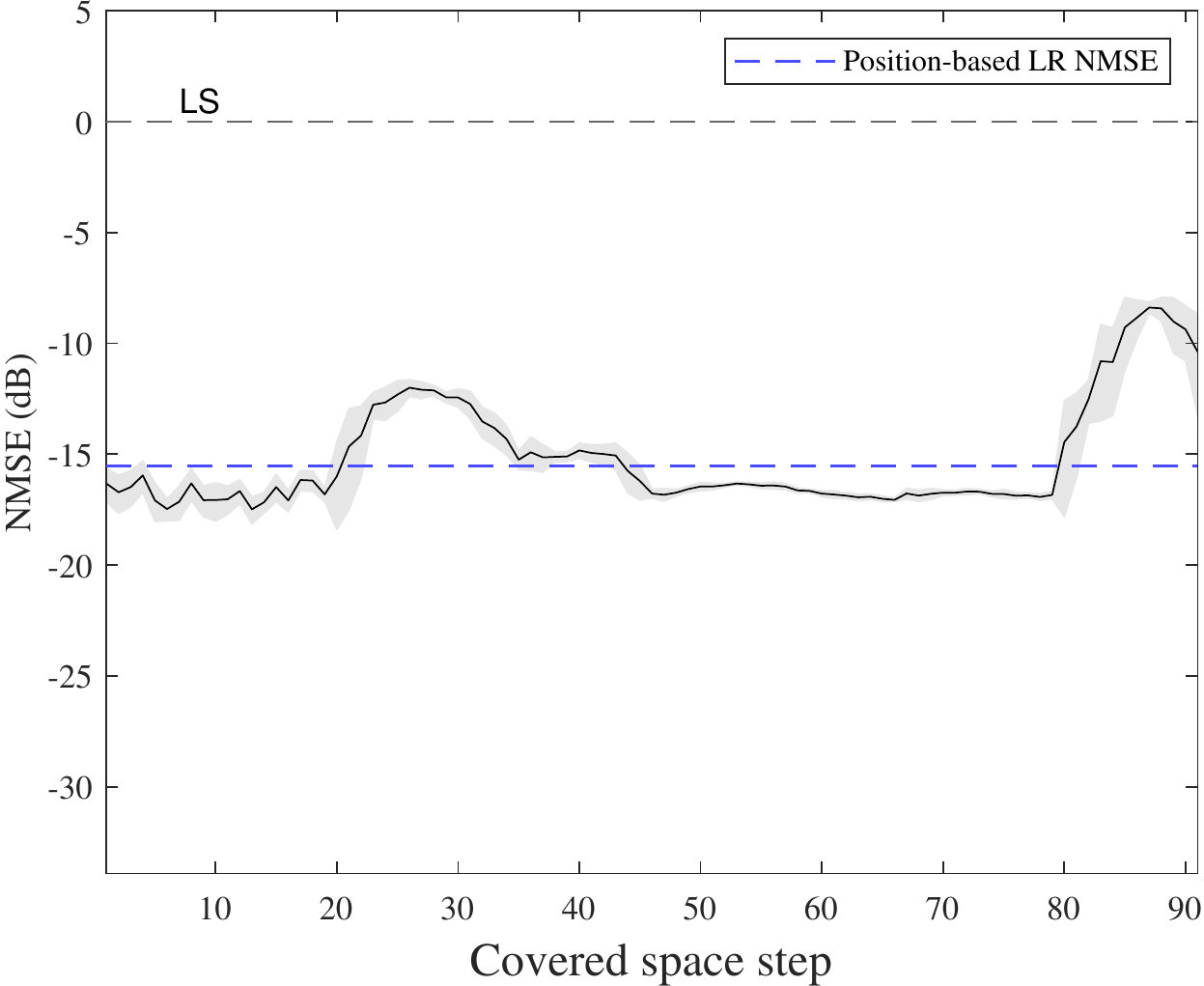}}\\
    \vspace{.5cm}
    \subfloat{\includegraphics[width=0.31\textwidth,valign=t]{images/results/maps/map_2.pdf}}
    \hspace{.5cm}
    \subfloat{\includegraphics[width=0.45\textwidth,valign=t]{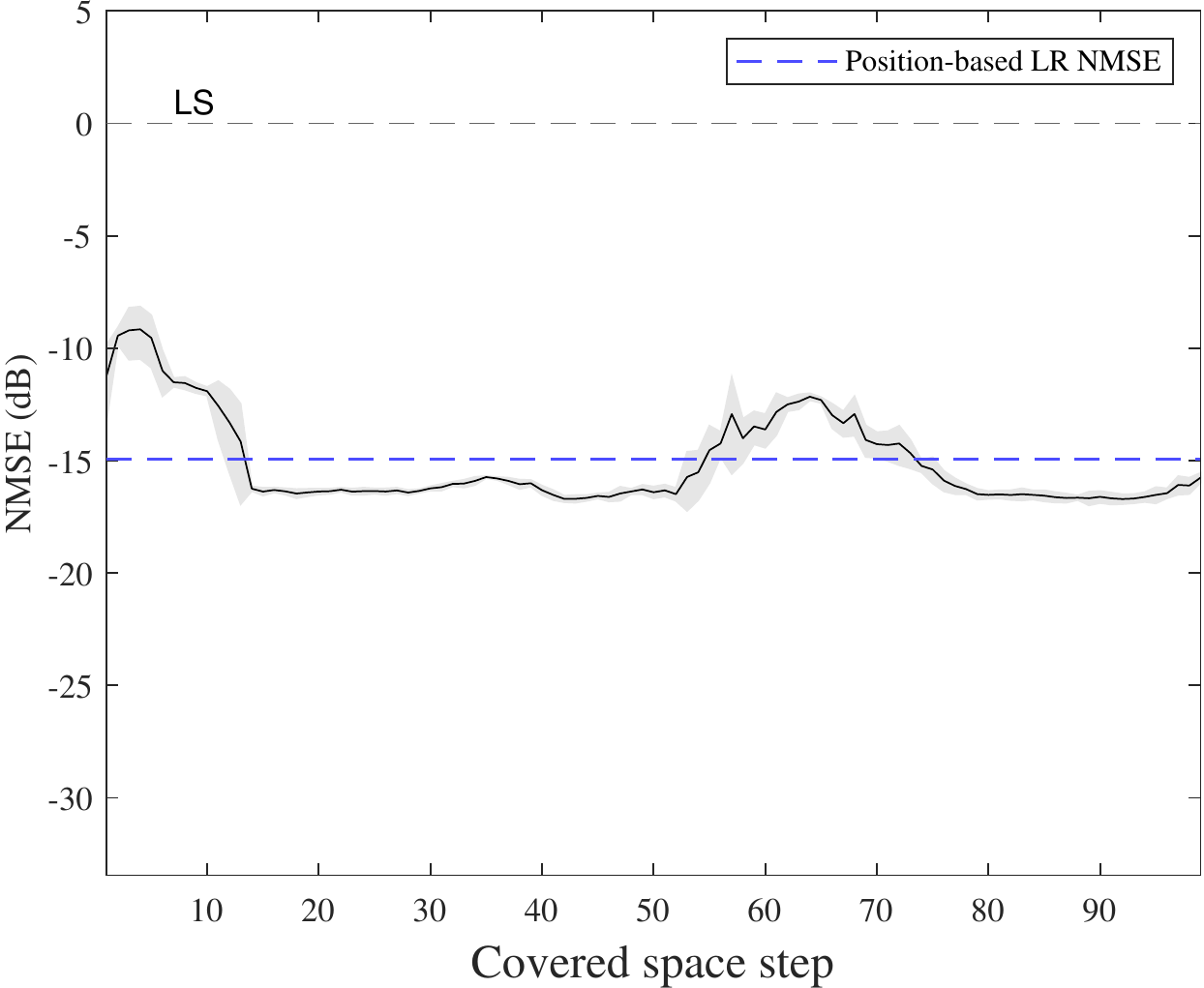}}
    \caption{Evaluation over sample trajectories in the training scenario a) for $B=50$ MHz.}
    \label{fig:results_over_sample_trajectories_50MHz}
\end{figure}

\begin{figure}
    \centering
    \subfloat[Direct model transfer\label{fig:box_plot_50MHz}]{\includegraphics[width=0.45\textwidth]{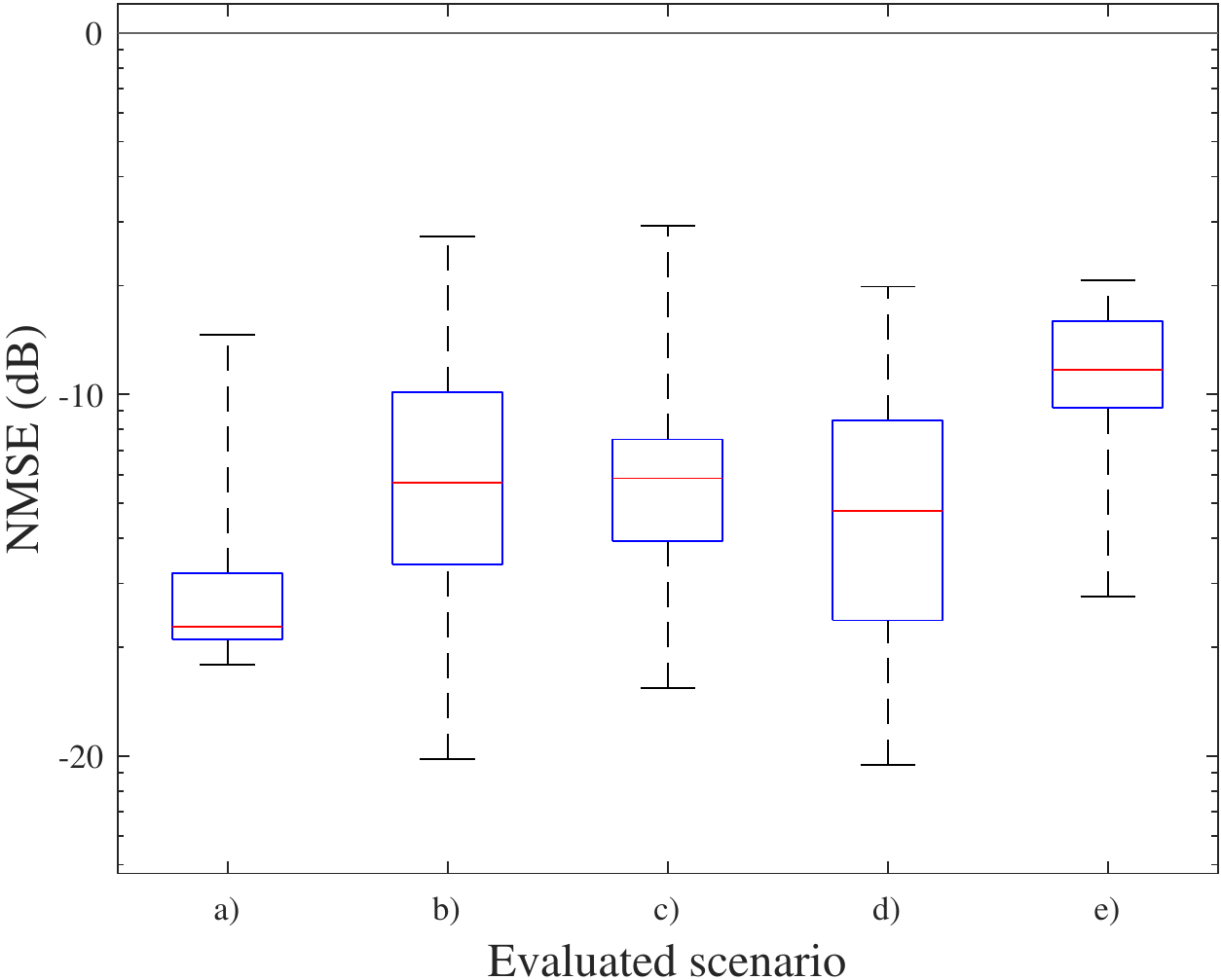}}
    \hspace{1.5cm}
    \subfloat[Model transfer with retraining of last two layers\label{fig:box_plot_50MHz_TL}]{\includegraphics[width=0.45\textwidth]{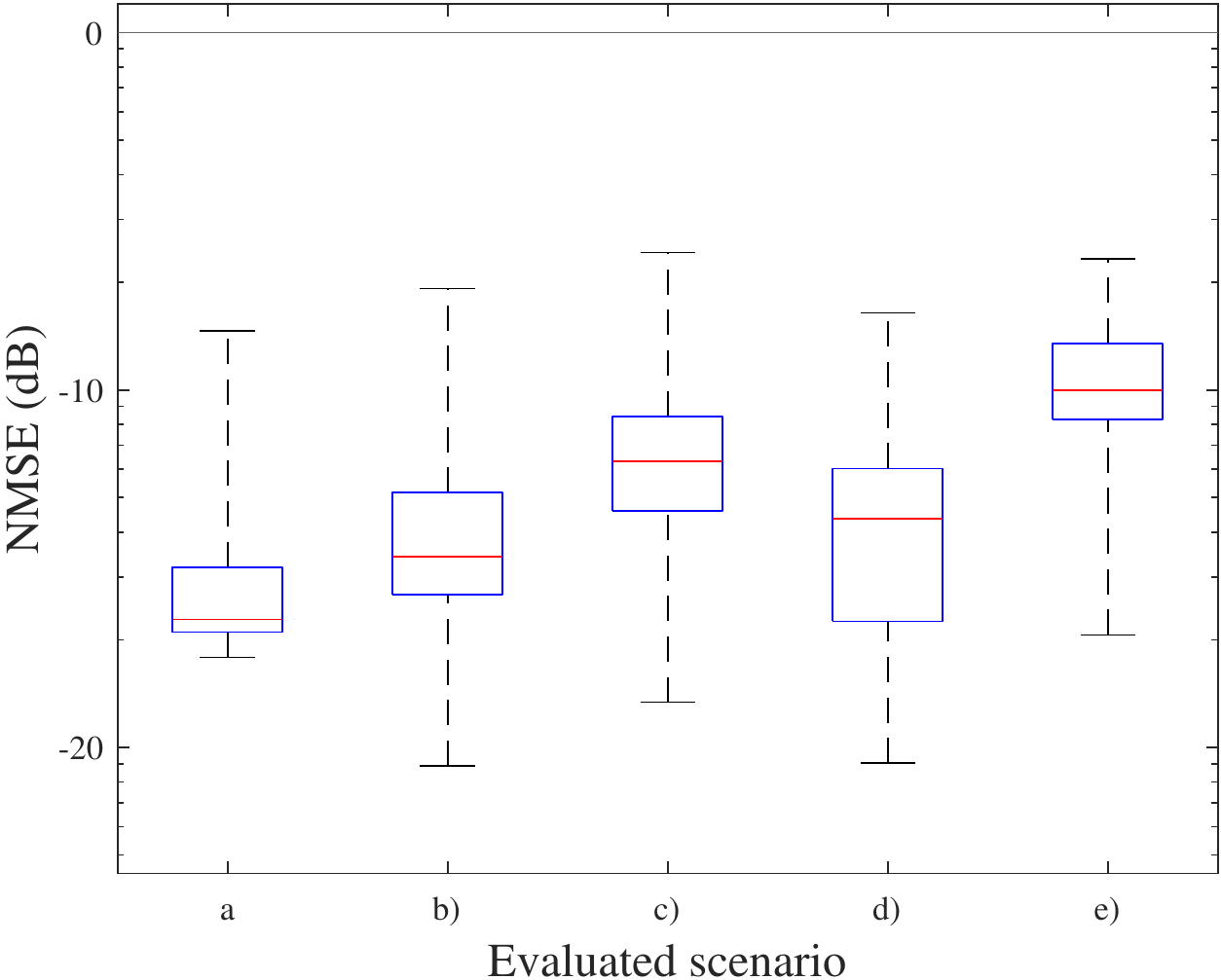}}
    \caption{Box plots showing median, interquartile range, and total range for the NMSE achieved using the DNN trained on scenario a) over the remaining scenarios (on the left), and the DNN trained on scenario a) with the last two fully-connected layers retrained on the specific scenario (on the right) for $B=50$ MHz.}
    \label{fig:three graphs}
\end{figure}

We show the results obtained evaluating the proposed DL-based channel estimation method to frequency-selective MIMO channels, i.e., $B=50$ MHz. After training the proposed model on the reference scenario a) in Fig. \ref{fig:scenarios_for_TL}, we analyse---as for the previously discussed frequency-flat case---its generalization to scenarios b), c), d), e). In this case, we first test the direct application of the trained model to the new urban scenarios, without any further retraining, having fixed the temporal channel length $W=22$ as the maximum over all the scenarios. We then examine whether any improvement can be obtained by retraining some network layers over LS and LR channel estimates proper of the specific application scenario.

\subsubsection{Performance of the DNN model on the reference urban scenario}
The considered DNN model has $3$ convolutional layers and $4$ fully-connected layers. The first two convolutional layers employ $64$ filters, while the third one uses a single filter. Differently from the frequency-flat condition, the three convolutional layers use $1 \times 3$ convolutional kernels, in order to jointly capture temporal features among consecutive temporal channel taps. The fully-connected layers are composed of $100$ neurons each. This leads to $\approx 4.7 \times 10^6$ trainable model parameters $\boldsymbol{\Theta}$. We train the DNN model on the reference scenario a), selecting the diversity orders $\bar{r}^{\mathrm{Tx}}_{\mathrm{S}} = 4$,  $\bar{r}^{\mathrm{Rx}}_{\mathrm{S}} = 8$, $\bar{r}_{\mathrm{T}} = 5$. With the considered setting, the DNN converges within $10$ training iterations to the average NMSE of the frequency-flat case, i.e., $-14.9$ dB, compared to the $-15.6$ dB obtained with the position-based LR method. Fig. \ref{fig:results_over_sample_trajectories_50MHz} shows the NMSE performance over the same 2 vehicular trajectories of Fig. \ref{fig:results_over_sample_trajectories_1MHz} (scenario a)). Although with more variability, even in the $B=50$ MHz case the DL model is able to provide comparable NMSE performance of the reference position-based LR method; as previously, a similar behavior is observed over the other 3 reference trajectories of a).

\subsubsection{Generalization of the DNN model to different urban scenarios}

To evaluate the generalization capabilities of the DL model in the frequency-selective channel case, we apply two different procedures: i) we directly test the model trained on reference scenario a) against scenarios b), c), d), and e), without any further re-training, and ii) starting from the model trained on scenario a), we fine-tune it by training only the last two fully-connected layers on the specific application scenario. We observed that the fine-tuning procedure converges after training the network with LS and LR channel estimates gathered in $10$ vehicle passages for each trajectory type in the target urban scenario.
Figures \ref{fig:box_plot_50MHz} and \ref{fig:box_plot_50MHz_TL} show the NMSE performance achieved over the evaluated scenarios respectively for procedures i) and ii) by means of box plots, where the red line represents the median, the box encloses the interval between the first and the third quartiles, and the outer bars delimit the range of observed NMSE performances. We notice that, compared to the frequency-flat case, the NMSE performance slightly deteriorate when transferring information to new scenarios, both in absence of retraining and with an explicit TL fine-tuning, still outperforming LS channel estimate by at least $10$ dB. This effect is a consequence of a greater variability of the MIMO channel eigenmodes due to the non-negligible temporal component ($W\gg1$). However, we did not observe any remarkable improvement applying a TL procedure, that only allows for a NMSE decrease of $1-2$ dB for scenarios b), d) and e).

\section{Conclusion}\label{sect:conclusion}

This paper addresses the problem of MIMO channel estimation in future 6G V2X systems proposing a novel DL-based LR channel estimation method. The proposed method leverages the received signal at the BS from road-induced recurrent vehicular UEs passages to design and train a DNN for the inference of MIMO channel eigenmodes. The goal is to improve conventional LS MIMO channel estimates without the need of any information on UEs' position. Exploiting the expressive power of DL and a training on LS and LR channel estimates collected over a whole radio cell, the proposed method requires only single input LS channel estimates to effectively infer the corresponding channel modes. Compared to a position-based LR channel estimation---which requires $L \approx 100$ pilot signals from as many vehicle passages for each location within a radio cell---this remarkably reduces any position-based training still achieving comparable NMSE performance.

Numerical results using realistic vehicular traffic and mmWave ray-tracing data show that the proposed DL-based LR method outperforms LS in terms of NMSE ($\approx 15$ dB) on channel estimation in both frequency-flat and frequency-selective channel cases, and attains the performance of the position-based LR, which in turn attains the theoretical MSE bound. Moreover, we show that the proposed DL model can be trained to infer the MIMO channel eigenmodes on a reference scenario, and then can be effectively transferred to urban scenarios (e.g., radio cells) characterized by substantially different space-time channel features, providing comparable NMSE performance without an explicit transfer learning fine-tuning procedure. This result allows to drastically reduce the number of training vehicles used to train the DNN, easing the practical implementation and motivating the application to future 6G V2X systems.

\section*{Acknowledgment}
The research has been carried out in the framework of the Joint Lab between Huawei and Politecnico di Milano.

\bibliographystyle{IEEEtran}
\bibliography{main}

\begin{thebibliography}{10}
\providecommand{\url}[1]{#1}
\csname url@samestyle\endcsname
\providecommand{\newblock}{\relax}
\providecommand{\bibinfo}[2]{#2}
\providecommand{\BIBentrySTDinterwordspacing}{\spaceskip=0pt\relax}
\providecommand{\BIBentryALTinterwordstretchfactor}{4}
\providecommand{\BIBentryALTinterwordspacing}{\spaceskip=\fontdimen2\font plus
\BIBentryALTinterwordstretchfactor\fontdimen3\font minus
  \fontdimen4\font\relax}
\providecommand{\BIBforeignlanguage}[2]{{%
\expandafter\ifx\csname l@#1\endcsname\relax
\typeout{** WARNING: IEEEtran.bst: No hyphenation pattern has been}%
\typeout{** loaded for the language `#1'. Using the pattern for}%
\typeout{** the default language instead.}%
\else
\language=\csname l@#1\endcsname
\fi
#2}}
\providecommand{\BIBdecl}{\relax}
\BIBdecl

\bibitem{Garcia2021_5GNRtutorial}
M.~H.~C. {Garcia}, A.~{Molina-Galan}, M.~{Boban}, J.~{Gozalvez},
  B.~{Coll-Perales}, T.~{Şahin}, and A.~{Kousaridas}, ``A tutorial on {5G NR
  V2X} communications,'' \emph{IEEE Communications Surveys Tutorials}, pp.
  1--1, 2021.

\bibitem{Wymeersch2021_6G}
C.~{De Lima}, D.~{Belot}, R.~{Berkvens}, A.~{Bourdoux}, D.~{Dardari},
  M.~{Guillaud}, M.~{Isomursu}, E.~S. {Lohan}, Y.~{Miao}, A.~N. {Barreto},
  M.~R.~K. {Aziz}, J.~{Saloranta}, T.~{Sanguanpuak}, H.~{Sarieddeen},
  G.~{Seco-Granados}, J.~{Suutala}, T.~{Svensson}, M.~{Valkama}, B.~{Van
  Liempd}, and H.~{Wymeersch}, ``Convergent communication, sensing and
  localization in {6G} systems: An overview of technologies, opportunities and
  challenges,'' \emph{IEEE Access}, vol.~9, pp. 26\,902--26\,925, 2021.

\bibitem{6834753}
M.~R. {Akdeniz}, Y.~{Liu}, M.~K. {Samimi}, S.~{Sun}, S.~{Rangan}, T.~S.
  {Rappaport}, and E.~{Erkip}, ``Millimeter wave channel modeling and cellular
  capacity evaluation,'' \emph{IEEE Journal on Selected Areas in
  Communications}, vol.~32, no.~6, pp. 1164--1179, 2014.

\bibitem{Akyildiz2015_subTHz}
C.~{Han}, A.~O. {Bicen}, and I.~F. {Akyildiz}, ``Multi-ray channel modeling and
  wideband characterization for wireless communications in the terahertz
  band,'' \emph{IEEE Transactions on Wireless Communications}, vol.~14, no.~5,
  pp. 2402--2412, 2015.

\bibitem{wang2018survey}
C.-X. Wang, J.~Bian, J.~Sun, W.~Zhang, and M.~Zhang, ``A survey of {5G} channel
  measurements and models,'' \emph{IEEE Communications Surveys \& Tutorials},
  vol.~20, no.~4, pp. 3142--3168, 2018.

\bibitem{kutty2015beamforming}
S.~Kutty and D.~Sen, ``Beamforming for millimeter wave communications: An
  inclusive survey,'' \emph{IEEE Communications Surveys \& Tutorials}, vol.~18,
  no.~2, pp. 949--973, 2015.

\bibitem{TS38213}
3GPP, ``{NR: Physical layer procedures for control},'' Third Generation
  Partnership Project (3GPP), Tech. Rep., 01 2020.

\bibitem{Guo2017_MUSIC_mmWChEst}
Z.~{Guo}, X.~{Wang}, and W.~{Heng}, ``Millimeter-wave channel estimation based
  on 2-{D} beamspace music method,'' \emph{IEEE Transactions on Wireless
  Communications}, vol.~16, no.~8, pp. 5384--5394, 2017.

\bibitem{Liao2017_ESPRIT_ChEst}
A.~{Liao}, Z.~{Gao}, Y.~{Wu}, H.~{Wang}, and M.~{Alouini}, ``{2D Unitary ESPRIT
  Based Super-Resolution Channel Estimation for Millimeter-Wave Massive MIMO
  With Hybrid Precoding},'' \emph{IEEE Access}, vol.~5, pp. 24\,747--24\,757,
  2017.

\bibitem{bajwa2010compressed}
W.~U. Bajwa, J.~Haupt, A.~M. Sayeed, and R.~Nowak, ``Compressed channel
  sensing: A new approach to estimating sparse multipath channels,''
  \emph{Proceedings of the IEEE}, vol.~98, no.~6, pp. 1058--1076, 2010.

\bibitem{Wang2018_CShardwareimpairments}
Y.~{Wu}, Y.~{Gu}, and Z.~{Wang}, ``{Channel Estimation for mmWave MIMO With
  Transmitter Hardware Impairments},'' \emph{IEEE Communications Letters},
  vol.~22, no.~2, pp. 320--323, 2018.

\bibitem{Nicoli2003}
M.~{Nicoli}, O.~{Simeone}, and U.~{Spagnolini}, ``Multislot estimation of
  fast-varying space-time communication channels,'' \emph{IEEE Transactions on
  Signal Processing}, vol.~51, no.~5, pp. 1184--1195, 2003.

\bibitem{Cerutti2020}
A.~{Brighente}, M.~{Cerutti}, M.~{Nicoli}, S.~{Tomasin}, and U.~{Spagnolini},
  ``{Estimation of Wideband Dynamic mmWave and THz Channels for 5G Systems and
  Beyond},'' \emph{IEEE Journal on Selected Areas in Communications}, vol.~38,
  no.~9, pp. 2026--2040, 2020.

\bibitem{mizmizi2021channel}
M.~Mizmizi, D.~Tagliaferri, D.~Badini, C.~Mazzucco, and U.~Spagnolini,
  ``{Channel Estimation for 6G V2X Hybrid Systems Using Multi-Vehicular
  Learning},'' \emph{IEEE Access}, vol.~9, pp. 95\,775--95\,790, 2021.

\bibitem{SalhDL6G2021}
A.~Salh, L.~Audah, N.~S.~M. Shah, A.~Alhammadi, Q.~Abdullah, Y.~H. Kim, S.~A.
  Al-Gailani, S.~A. Hamzah, B.~A.~F. Esmail, and A.~A. Almohammedi, ``{A Survey
  on Deep Learning for Ultra-Reliable and Low-Latency Communications Challenges
  on 6G Wireless Systems},'' \emph{IEEE Access}, vol.~9, pp. 55\,098--55\,131,
  2021.

\bibitem{osheaIntroductionDeepLearning2017a}
T.~O'Shea and J.~Hoydis, ``An {{Introduction}} to {{Deep Learning}} for the
  {{Physical Layer}},'' \emph{IEEE Transactions on Cognitive Communications and
  Networking}, vol.~3, no.~4, pp. 563--575, Dec. 2017.

\bibitem{zhangDeepLearningMobile2019}
C.~Zhang, P.~Patras, and H.~Haddadi, ``Deep {{Learning}} in {{Mobile}} and
  {{Wireless Networking}}: {{A Survey}},'' \emph{IEEE Communications Surveys
  Tutorials}, vol.~21, no.~3, pp. 2224--2287, thirdquarter 2019.

\bibitem{huangDeepLearningPhysicalLayer2020}
H.~Huang, S.~Guo, G.~Gui, Z.~Yang, J.~Zhang, H.~Sari, and F.~Adachi, ``Deep
  {{Learning}} for {{Physical}}-{{Layer 5G Wireless Techniques}}:
  {{Opportunities}}, {{Challenges}} and {{Solutions}},'' \emph{IEEE Wireless
  Communications}, vol.~27, no.~1, pp. 214--222, Feb. 2020.

\bibitem{wen2018deep}
C.-K. Wen, W.-T. Shih, and S.~Jin, ``{Deep learning for massive MIMO CSI
  feedback},'' \emph{IEEE Wireless Communications Letters}, vol.~7, no.~5, pp.
  748--751, 2018.

\bibitem{liao2019deep}
Y.~Liao, Y.~Hua, and Y.~Cai, ``Deep learning based channel estimation algorithm
  for fast time-varying mimo-ofdm systems,'' \emph{IEEE Communications
  Letters}, 2019.

\bibitem{ulyanov2018deep}
D.~Ulyanov, A.~Vedaldi, and V.~Lempitsky, ``Deep image prior,'' in
  \emph{Proceedings of the IEEE conference on computer vision and pattern
  recognition}, 2018, pp. 9446--9454.

\bibitem{balevi2020massive}
E.~Balevi, A.~Doshi, and J.~G. Andrews, ``Massive mimo channel estimation with
  an untrained deep neural network,'' \emph{IEEE Transactions on Wireless
  Communications}, vol.~19, no.~3, pp. 2079--2090, 2020.

\bibitem{nguyen2021transfer}
C.~T. Nguyen, N.~Van~Huynh, N.~H. Chu, Y.~M. Saputra, D.~T. Hoang, D.~N.
  Nguyen, Q.-V. Pham, D.~Niyato, E.~Dutkiewicz, and W.-J. Hwang, ``Transfer
  learning for future wireless networks: A comprehensive survey,'' \emph{arXiv
  preprint arXiv:2102.07572}, 2021.

\bibitem{alves2021deep}
W.~Alves, I.~Correa, N.~Gonz{\'a}lez-Prelcic, and A.~Klautau, ``Deep transfer
  learning for site-specific channel estimation in low-resolution mmwave
  mimo,'' \emph{IEEE Wireless Communications Letters}, 2021.

\bibitem{yang2020deep}
Y.~Yang, F.~Gao, Z.~Zhong, B.~Ai, and A.~Alkhateeb, ``{Deep transfer
  learning-based downlink channel prediction for FDD massive MIMO systems},''
  \emph{IEEE Transactions on Communications}, vol.~68, no.~12, pp. 7485--7497,
  2020.

\bibitem{altairwinprop}
``{Altair WinProp},''
  \url{https://altairhyperworks.com/product/feko/winprop-propagation-modeling}.

\bibitem{SUMO2018}
\BIBentryALTinterwordspacing
P.~A. Lopez, M.~Behrisch, L.~Bieker-Walz, J.~Erdmann, Y.-P. Fl{\"o}tter{\"o}d,
  R.~Hilbrich, L.~L{\"u}cken, J.~Rummel, P.~Wagner, and E.~Wie{\ss}ner,
  ``{Microscopic Traffic Simulation using SUMO},'' in \emph{The 21st IEEE
  International Conference on Intelligent Transportation Systems}.\hskip 1em
  plus 0.5em minus 0.4em\relax IEEE, 2018. [Online]. Available:
  \url{https://elib.dlr.de/124092/}
\BIBentrySTDinterwordspacing

\bibitem{cazzella2021positionagnostic}
L.~Cazzella, D.~Tagliaferri, M.~Mizmizi, M.~Matteucci, D.~Badini, C.~Mazzucco,
  and U.~Spagnolini, ``{Position-agnostic Algebraic Estimation of 6G V2X MIMO
  Channels via Unsupervised Learning},'' 2021.

\bibitem{8690640}
M.~{Mizmizi}, S.~{Mandelli}, S.~{Saur}, and L.~{Reggiani}, ``Robust and
  flexible tracking of vehicles exploiting soft map-matching and data fusion,''
  in \emph{2018 IEEE 88th Vehicular Technology Conference (VTC-Fall)}, 2018,
  pp. 1--5.

\bibitem{maas2013rectifier}
A.~L. Maas, A.~Y. Hannun, A.~Y. Ng \emph{et~al.}, ``Rectifier nonlinearities
  improve neural network acoustic models,'' in \emph{Proc. icml}, vol.~30,
  no.~1.\hskip 1em plus 0.5em minus 0.4em\relax Citeseer, 2013, p.~3.

\bibitem{ioffe2015batch}
S.~Ioffe and C.~Szegedy, ``Batch normalization: Accelerating deep network
  training by reducing internal covariate shift,'' in \emph{International
  conference on machine learning}.\hskip 1em plus 0.5em minus 0.4em\relax PMLR,
  2015, pp. 448--456.

\bibitem{golub2012matrix}
G.~H. Golub and C.~F. Van~Loan, \emph{Matrix computations}.\hskip 1em plus
  0.5em minus 0.4em\relax JHU press, 2012, vol.~3.

\bibitem{wan2019automatic}
Z.-Q. Wan and S.-X. Zhang, ``Automatic differentiation for complex valued
  svd,'' \emph{arXiv preprint arXiv:1909.02659}, 2019.

\bibitem{kingma2014adam}
D.~P. Kingma and J.~Ba, ``Adam: A method for stochastic optimization,''
  \emph{arXiv preprint arXiv:1412.6980}, 2014.

\bibitem{absil2009optimization}
P.-A. Absil, R.~Mahony, and R.~Sepulchre, \emph{Optimization algorithms on
  matrix manifolds}.\hskip 1em plus 0.5em minus 0.4em\relax Princeton
  University Press, 2009.

\end{thebibliography}

\end{document}